\documentclass[a4paper]{article}

\usepackage{lineno}
\renewcommand{\linenumbers}{}

\usepackage{url}
\usepackage[inline]{enumitem}
\usepackage{authblk}

\usepackage{longtable}
\usepackage{amsmath}
\usepackage{url}
\usepackage[colorlinks,bookmarksnumbered=true,pagebackref=true]{hyperref}
\hypersetup{
        bookmarks=true,
        bookmarksnumbered=true,
        pdfstartview={FitH},
        citecolor={blue},
        linkcolor={blue},
        urlcolor={black},
        pdfpagemode={UseOutlines}
}
\makeatletter
\newcommand\org@hypertarget{}
\let\org@hypertarget\hypertarget
\renewcommand\hypertarget[2]{\Hy@raisedlink{\org@hypertarget{#1}{}}#2}
\makeatother
\usepackage[boxed,linesnumbered]{algorithm2e}
\usepackage{braket}
\usepackage{xspace}
\usepackage{subfig}
\usepackage{soul}
\usepackage{numprint}
\usepackage{tikz,pgf,pgfplots,pgfplotstable}
\usetikzlibrary{arrows,decorations,backgrounds,matrix,automata,
trees,shapes,shadows,plotmarks,calc,positioning,patterns,chains,fit}
\usepgfplotslibrary{groupplots,colorbrewer,patchplots}
\pgfplotsset{compat=1.16}

\newcommand{\Omit}[1]{}

\newcommand{\card}[1]{\lvert #1 \rvert}
\newcommand\A{\mathcal{A}}
\newcommand\B{\mathcal{B}}
\newcommand\D{\mathcal{D}}

\newcommand\I{\mathcal{I}}
\newcommand\M{\mathcal{M}}

\newcommand\R{\mathcal{R}}
\newcommand\V{\mathcal{V}}

\newcommand\CE{\texttt{Considered Evacuator}\xspace}
\newcommand\CG{\texttt{Community Guided}\xspace}
\newcommand\TD{\texttt{Threat Denier}\xspace}
\newcommand\WW{\texttt{Worried Waverer}\xspace}
\newcommand\RD{\texttt{Responsibility Denier}\xspace}
\newcommand\DE{\texttt{Dependent Evacuator}\xspace}
\newcommand\EI{\texttt{Experienced Independent}\xspace}
\newcommand\UT{\texttt{Unknown Type}\xspace}

\newcommand\Advice{\texttt{Advice}\xspace}
\newcommand\WatchAndAct{\texttt{WatchAndAct}\xspace}
\newcommand\EmergencyWarning{\texttt{EmergencyWarning}\xspace}
\newcommand\EvacuateNow{\texttt{EvacuateNow}\xspace}
\newcommand\VisibleSmoke{\texttt{VisibleSmoke}\xspace}
\newcommand\VisibleEmbers{\texttt{VisibleEmbers}\xspace}
\newcommand\VisibleFire{\texttt{VisibleFire}\xspace}
\newcommand\ThresholdInitial{\texttt{ThresholdInitial}\xspace}
\newcommand\ThresholdFinal{\texttt{ThresholdFinal}\xspace}
 
\pgfrealjobname{main}

\title{What will they do?\\Modelling self-evacuation archetypes}

\author[1,5]{Dhirendra Singh}
\author[2]{Ken Strahan}
\author[3]{Jim McLennan}
\author[4]{Joel Robertson}
\author[1]{Bhagya Wickramasinghe}

\affil[1]{School of Computing Technologies, RMIT University\authorcr GPO Box 2476, Melbourne VIC 3001, Australia}
\affil[2]{Strahan Research,
Level 3, 90 Williams St, Melbourne VIC 3000, Australia}
\affil[3]{School of Psychology and Public Health,
La Trobe University\authorcr Bundoora VIC 3086, Australia}
\affil[4]{School of Science,
RMIT University\authorcr GPO Box 2476, Melbourne VIC 3001, Australia}
\affil[5]{CSIRO Data61,
Goods Shed, Village Street, Docklands VIC 3008, Australia}

\begin{document}

\maketitle

\begin{abstract}
A decade on from the devastating Black Saturday bushfires in Victoria,
Australia, we are at a point where computer simulations of community evacuations
are starting to be used within the emergency services. While fire progression
modelling is embedded in strategic and operational settings at all levels of
government across Victoria, modelling of community response to such fires is
only just starting to be evaluated in earnest. For community response models to
become integral to bushfire planning and preparedness, the key question to be
addressed is: when faced with a bushfire, what will a community really do?
Typically this understanding has come from local experience and expertise within
the community and services, however the trend is to move towards more informed
data driven approaches. In this paper we report on the latest work within the
emergency sector in this space. Particularly, we discuss the
application of Strahan et al.’s self-evacuation \textit{archetypes} to an
agent-based model of community evacuation in regional Victoria. This work is
part of the consolidated bushfire evacuation modelling collaboration between
several emergency management stakeholders.
 \end{abstract}

\linenumbers

\section{Introduction}\label{sec:introduction}

Wildfires (called \textit{bushfires} in Australia) now pose serious threats to
life and property in numerous countries globally~\footnote{For a global list of
serious wildfires see \url{https://en.wikipedia.org/wiki/List_of_wildfires}}.
There is general agreement among environmental scientists and wildfire
researchers that severe wildfires will become more frequent in future, in many
regions, as a result of climate change~\cite{tedim_defining_2018}. The immediate
threats to human life from wildfire result from the intense heat generated by
large masses of burning vegetation fuels. Most wildfire fatalities are caused by
radiant or convective heat; vehicle accidents during late evacuation have also
caused significant numbers of civilian
casualties~\cite{blanchi_environmental_2014}. Residents threatened by a severe
wildfire have, in principle, three protective action options: contesting the
threat by actively defending their property, enduring the threat by sheltering
passively on their property, or avoiding the threat by timely evacuation to a
safer location—which may be remote from the threatened area (e.g., at a
designated evacuation centre) or nearby in an ad hoc last-resort shelter
location~\cite{blanchi_surviving_2018}. Timely evacuation away from the
threatened location is the protective action preferred overwhelmingly by
authorities in countries where wildfires are a serious hazard. In North America
and Europe evacuation is usually mandated; in Australia bushfire evacuation is
typically advisory, forced evacuations are rare.

Historically, two research perspectives have dominated hazard evacuation
research generally, including wildfire threat evacuation: an engineering
perspective, and a behavioural perspective~\cite{cova_protective_2009}. An
engineering perspective focusses on vehicular (mostly) traffic flow
modelling—volume and timing analysis. A behavioural perspective focusses on
understanding factors influencing public compliance with evacuation orders or
advice— preparation and readiness. More recently, there has been an integration
of these two perspectives so that understandings of evacuation behaviour are
incorporated into modelling in order to improve the accuracy of evacuation
traffic routing and time
estimates~\cite{cova_protective_2009,lovreglio_modelling_2019}. A recent review
of research findings about behavioural factors in evacuation under wildfire
threat~\cite{mclennan_should_2019} concluded the following (pp. 503):
\begin{enumerate*}[label=(\roman*)]

\item	Regardless of whether an evacuation is mandatory or advisory, some
residents will want to remain on their property to protect it.

\item	A significant percentage of residents will want to delay leaving because
they will not leave until they are certain that leaving is necessary.

\item Some residents who are not on their property when the evacuation warning
is given will attempt to return to their property first, rather than proceeding
to a safer location immediately.

\item Mandatory evacuation orders are likely to result in higher rates of
compliance and fewer residents delaying evacuation.

\item Residents of farms and other agribusinesses are more likely to stay and
defend their properties. Residents of amenity dwellings are more likely to
leave.

\item	Three important determinants of residents’ actions are:
\begin{enumerate*}[label=(\alph*)]
\item their prior plan about what they will do if threatened,
\item their perceived danger from the fire, and
\item their level of logistical preparations and readiness to evacuate.
\end{enumerate*}

\item Evacuation will be more difficult and involve longer delays for families
with:
\begin{enumerate*}[label=(\alph*)]
\item children,
\item members who have disabilities or special needs,
\item companion animals and/or livestock.
\end{enumerate*}

\end{enumerate*}

Safe evacuation is not simply a matter of authorities convincing residents to
leave early, sufficient means of safe egress must be available for the traffic
volume under the conditions of the day. Modelling the impact of a wildfire
threat on a specific location or area can assist authorities to better plan and
prepare for an evacuation under a range of vegetation fuel loads, fire weather
conditions, population characteristics, and egress capacities.
 
Models of bushfire progression~\cite{tolhurst_assessing_2011,denzer_spark_2015}
and traffic evacuation
dynamics~\cite{singh_emergency_2017,horni_making_2016,diaz_bdi_2016}  assist in
understanding the physical and social dynamics of bushfire threats.
However the integration of
householder decision response to imminent bushfire threat into dynamic travel
demand models of vehicular traffic is in its early
stages~\cite{russo_integrated_2014}.
While several studies that consider household behaviours in bushfires have been
conducted~\cite{mclennan_predictors_2014,mclennan_wait_2012,mclennan_householder_2012},
a failure to incorporate these important elements of human behaviour in computer
models continues to be a shortcoming~\cite{folk_provisional_2019}.
The infancy of this work may be explained by the lack of a systematic modelling
framework of household perceptions and response to
bushfire~\cite{russo_integrated_2014,dixit_modeling_2012}. Progress on
identifying functional requirements of a model~\cite{ronchi_e-sanctuary:_2017},
the factors to include~\cite{folk_provisional_2019} and recently, a means of
connecting physical conditions with householder decision-making and their
protective response are being made~\cite{lovreglio_modelling_2019}.

In this context an agent-based model~\cite{abar_agent_2017} of householder
decision-making, using the behavioural insights provided by the self-evacuation
archetypes~\cite{strahan_self-evacuation_2018,strahan_predicting_2019} was
developed. Factors affecting householder decision-making and how they respond to
different information and cues were systematically incorporated into a dynamic
vehicle travel demand model.
The archetype-based model enables perception and response-based predictions of
householders’ decision to stay or leave and the timing of their action.
Communities are modelled at the level of an individual who are each assigned to
an archetypal persona. Synthetic individuals in the simulation make decisions
based on their personal circumstance (such as their location and whether they
have dependants to attend to), their understanding of the unfolding bushfire
situation (built from what they ``hear'' from the emergency services and ``see''
based on their proximity to the fire), and their individual propensity for risk
(which differs between archetypes and also individuals of the same archetype).

The archetype-based behaviour model has been integrated into the Emergency
Evacuation Simulator (EES)~\cite{singh_emergency_2017} that is currently being
used by Victorian emergency services for understanding community response in
bushfire evacuation scenarios.
The EES allows community evacuation response to be simulated for any bushfire
situation and weather conditions.
It takes as input a time-varying fire progression shape output by the
Phoenix RapidFire model~\cite{tolhurst_assessing_2011}\footnote{Phoenix
RapidFire outputs for the chosen weather conditions are supplied by the
Victorian emergency services.}, and simulates the response of the affected
communities in terms of individuals' decisions to stay or leave, as well as
vehicular movements where decisions to leave have been made.
The situation-based cognitive decision making is modelled using the
Belief-Desire-Intention paradigm~\cite{rao_modeling_1991,cohen_intention_1990},
while the vehicular movements of the associated individual on the road network
are modelled using the MATSim traffic simulator~\cite{horni_multi-agent_2016},
with the coupling between the two provided by the BDI-MATSim integration
framework~\cite{singh_integrating_2016}.

Outputs of the simulation include full details of vehicles' trajectories along
with individuals' time-stamped decisions, which together constitute virtual
accounts of archetypal individuals' bushfire experience, and resemble anecdotal
accounts from survivors of real bushfires.
These outputs can be utilised by the emergency services in different ways.
Visualisations of the simulation can help to identify likely congestion
bottleneck areas in the road network. Outputs from a range of bushfire
situations can help improve understanding of likely, worst, and best case
evacuation outcomes for communities and inform planning, preparedness, and
community education programs. Analysis from simulation runs for current weather
conditions and possible bushfire scenarios can be used in an operational setting
on the day. A range of response strategies can be virtually tested, including
the timing and level of emergency messages sent to residents, as well as traffic
control points and road blockages that could be deployed to direct evacuating
vehicles readily towards safety.
 \section{Self-evacuation Archetypes}
\label{sec:archetypes}

An archetype is a typical character who resonates with an observer due to their
universally shared, fundamental characteristics of humanity based on myths,
legends and esoteric teachings~\cite{jung_archetypes_1968}.  Social cues,
replicated through dominant discourse~\cite{campbell_power_1988} and collective
memory, as shared experiences that are constructed and validated through social
interaction \cite{halbwachs_collective_1992} are also a basis for the formation
of individual and societal perceptions of archetypes.

Archetypes do not represent the circumstances or behaviour of individuals but
reflect typical groupings. Archetypes provide a framework for addressing the
questions about who does what, why and when they do it, reflecting individual
attitudes, needs, motivations, and issues fundamental to them. They provide
insights into how people think, feel, and act in a prevailing situation,
clarifying patterns of behaviour and drivers to action. Different archetypes
reflect a range of typical and generally predictable patterns of actions or
response which are influenced by the interaction between the context and a
variety of other salient factors. Archetypes also enable a greater understanding
of both the potential barriers to and the opportunities and strategies for
communicating and engaging with individuals.

Archetypes in a bushfire context are characterised by the way they understand
bushfire risk, and their attitudes, intentions and priorities including
self-efficacy and responsibility, bushfire experience, threat perception,
preparedness, use of environmental and social cues and networks, and intended
protective response.

\begin{longtable}{|p{0.2\columnwidth}p{0.35\columnwidth}p{0.35\columnwidth}|}
\caption{Self-evacuation archetypes of Strahan et
al.~\cite{strahan_self-evacuation_2018} (cf. Table~1).}\label{tab:archetypes}\\
\hline
{\bf Archetype} & {\bf Key Characteristics} & {\bf Evacuate or Remain}\\
\hline\hline
\RD	& Believe they are not responsible for their personal safety or for their
property	& Highly committed evacuators but expect others to direct and
assist\\\hline
\DE & Expect the emergency services to protect them and their property because
they are incapable of taking responsibility for themselves & Highly committed
evacuators but expect others to direct and assist\\\hline
\CE	& Having carefully considered evacuation, are committed to it as soon as
they are aware of a bushfire threat & Committed to self-directed
evacuation\\\hline
\CG & Seek guidance from neighbours, media and members of the community who they
see as knowledgeable, well informed and providing reliable advice &	Committed to
evacuation on community advice\\\hline
\WW & Prepare and equip their property and train to defend it but worry they
lack practical experience to fight bushfire putting their personal safety at
risk & Wavering between evacuating and remaining\\\hline
\TD	& Do not believe that their personal safety or property is threatened by
bushfire & Committed to remain at their property as perceived lack of threat
makes evacuation unnecessary\\\hline
\EI & Are highly knowledge, competent and experienced and are responsible and
self-reliant fighting bushfire & Highly committed to remaining to defend their
property because they are highly experienced and well prepared\\\hline
\end{longtable}

Seven bushfire self-evacuation archetypes reported by Strahan et
al.~\cite{strahan_self-evacuation_2018} in Table~\ref{tab:archetypes}
characterise the diverse attitudes and behaviours of typical groupings of
householders faced with making a protective decision during a bushfire.
The archetypes were established using the following process (cf.~\cite{strahan_self-evacuation_2018}).
Four hundred and fifty-seven householders who had recently experienced bushfire
participated in a telephone interview as part of a study of factors predicting
self-evacuation~\cite{strahan_predicting_2019}. The Protective Action Decision Model
(PADM)~\cite{lindell_protective_2012,strahan_protective_2018}
provided a theoretical framework for the survey questions, data from many of
which were converted to Z scores and analysed using the K-means cluster
procedure of IBM SPSS 24 to identify groupings of cases. The procedure was repeated for 3 to 8 clusters with 7
clusters (n= 31-93) achieved after 13 iterations and providing the greatest
stability. Univariate ANOVAs indicating clustered groups differed significantly
(p<0.05) on all except four variables. The seven clusters were subject to
explanatory discriminant function analysis to demonstrate
their statistical validity. Weighted linear combinations of variables that best
differentiated the clusters produced discriminant functions that accounted for a
statistically significant percentage of between group differences. 93.4\% of the
grouped cases were correctly classified and 79\% using a leave-one-out cross
validation approach.
 \section{Method}\label{sec:method}

We describe here the process of constructing a synthetic population of
archetypes for a new region, assigning evacuation behaviours to it, and putting
together and simulating a full bushfire evacuation scenario.

\subsection{Estimating the likelihood of archetypes}
\label{subsec:archetypes-probabilities}

We used Strahan et al.'s~\cite{strahan_self-evacuation_2018} original data
from telephone interviews of individuals (n=457) of bushfire affected
areas in the Perth (2014) and Adelaide Hills (2015) to extract the makeup of the
archetypes by demography.
The output of this step was a matrix $\M$ giving the likelihood of an individual
with the demographic \textit{signature} \texttt{<age,gender,hhtype>}, i.e., of a
given age, gender, and household type (rows), belonging to a certain archetype
(columns). A sample row of the matrix is shown in
Table~\ref{tab:strahan-sample}.
Some cells in $\M$ were evidently empty, due to a lack of samples for that
demographic signature in the original data. Where rows had only some missing
values, the empty cells were assigned a zero probability~\footnote{This step
could be improved by estimating the values of the empty cells based on
neighbouring cells in the matrix, though we did not do this here.}, such as for
the WW and DE cells in Table~\ref{tab:strahan-sample}. For rows where all cells
were empty, i.e., no sample existed for that type of person,  the signature was
allocated to a new \UT class with 100\% probability.
The size of the probabilities matrix was $\card{\M}=70\times8$, given by $8$
archetype columns (7 archetypes plus \UT), and $70$
signatures (7 age groups: $18-24$, $25-34$, $35-44$, $45-54$, $55-64$, $65-74$,
$75+$; 2 genders: Male, Female; and 5 household types: Single adults
with/without dependant children, Couples with/without dependent children, and
Group adults).

\begin{table}[!t]
  \centering
  \caption{Sample row from Strahan et al.'s~\cite{strahan_self-evacuation_2018}
  data (n=457) matrix $\M$, showing the likelihood (\%) of a 25-34 year
  old female living with a partner and children belonging to each archetype.
  The column names stand for CE:~\CE, CG:~\CG, TD:~\TD, WW:~\WW, RD:~\RD,
  DE~\DE, EI:~\EI, and UT:~\UT.}
  \label{tab:strahan-sample}
  \begin{tabular}{|p{0.44\columnwidth}|l@{\hspace{3pt}}l@{\hspace{3pt}}l@{\hspace{3pt}}l@{\hspace{3pt}}l@{\hspace{3pt}}l@{\hspace{3pt}}l@{\hspace{3pt}}l|}
    \hline
    Person & CE & CG & TD & WW & RD & DE & EI & UT\\
    \hline\hline
    Female, 25-34, Couple with dependants living at home
    & 2.4 & 4.8 & 2.4 & - & 2.4 & - & 2.4 & -\\
    \hline
  \end{tabular}
\end{table}

\begin{figure*}[!t]
  \centering
  \subfloat[][Distribution of archetypes by gender] {
    \includegraphics[width=0.8\columnwidth]{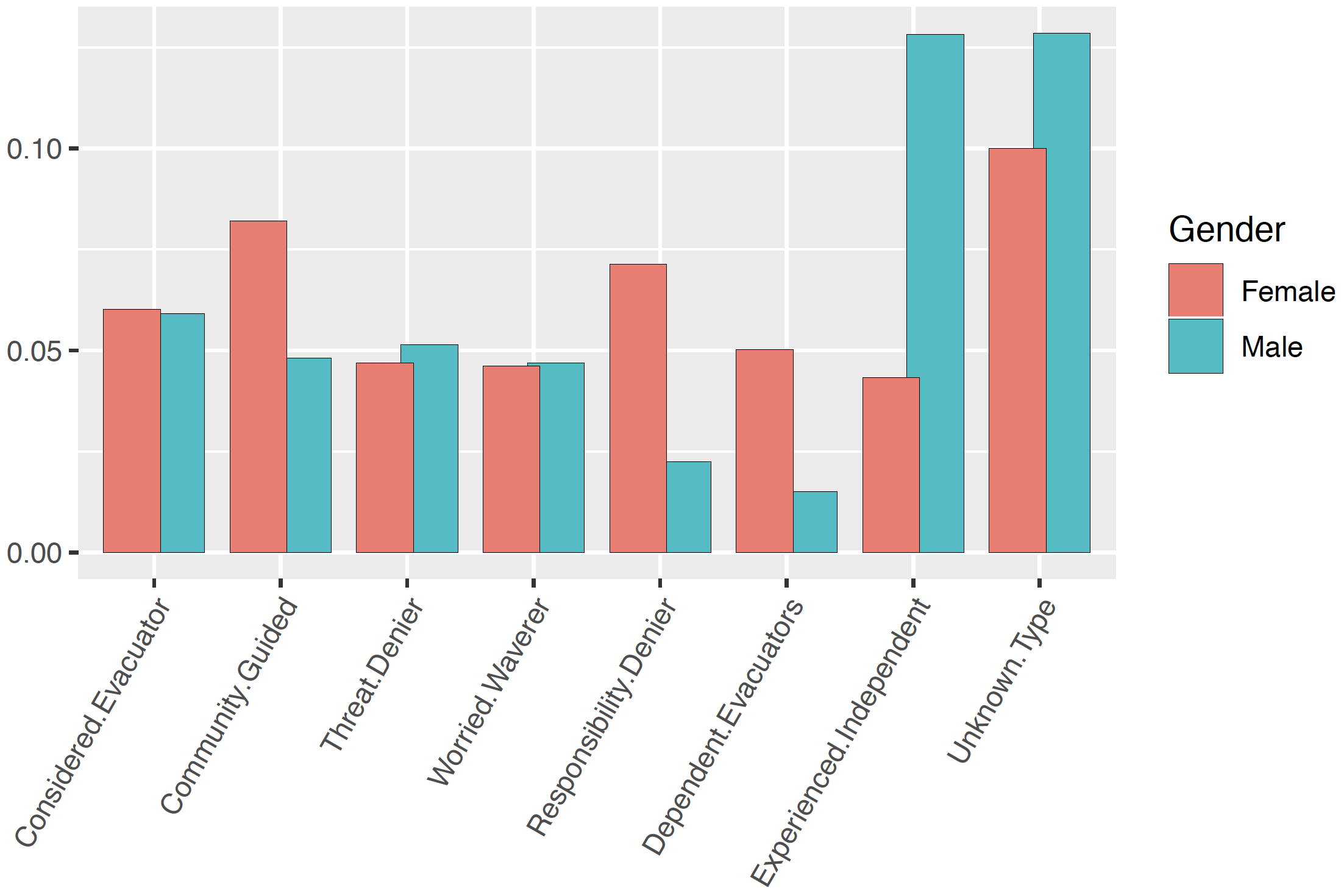}
    \label{fig:archetypes-baseline-gender}
    }
  \\
  \subfloat[][Distribution of archetypes by age groups] {
    \includegraphics[width=0.8\columnwidth]{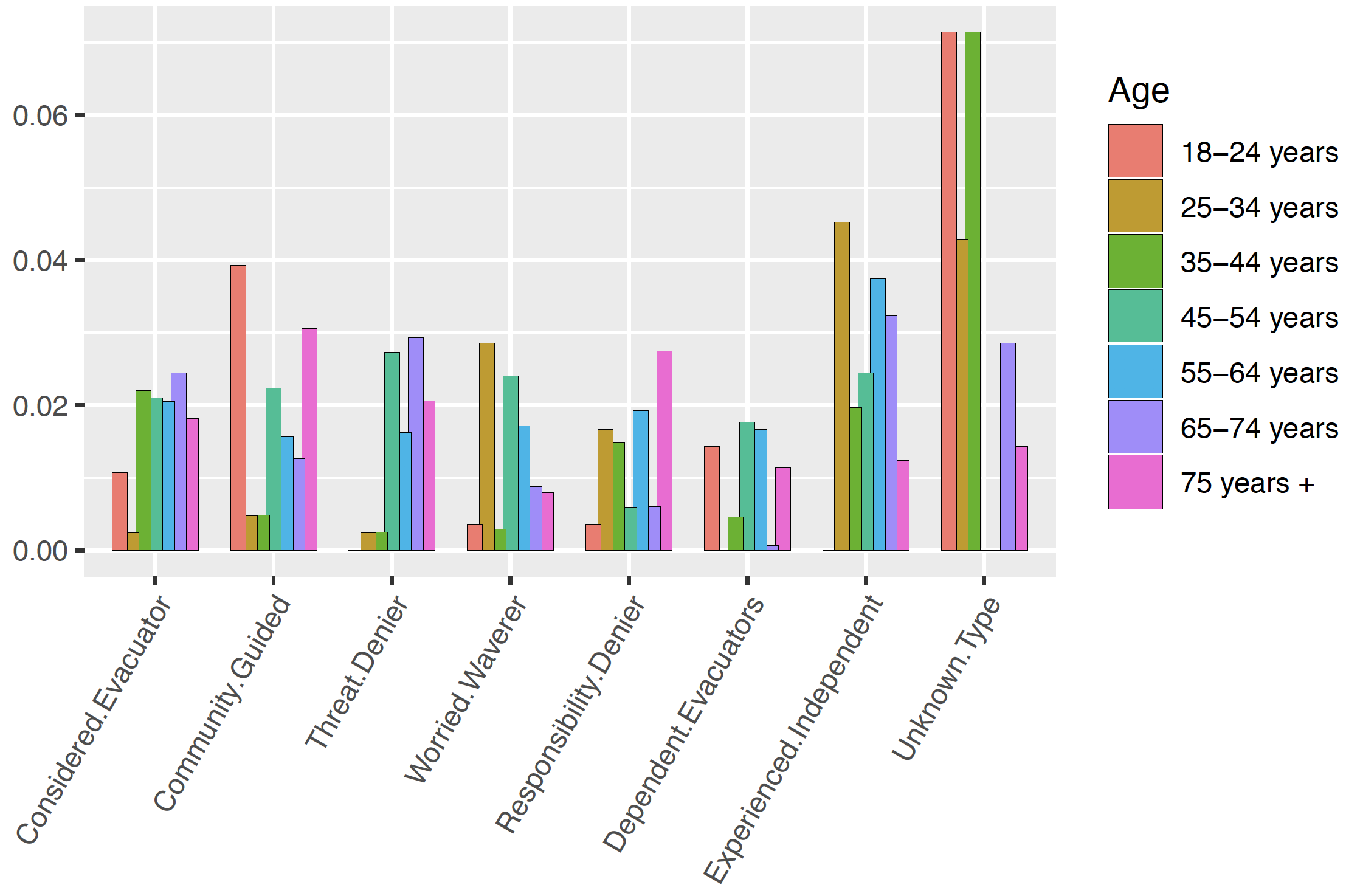}
    \label{fig:archetypes-baseline-age}
  }
  \caption{Calculated distribution of archetypes in Strahan et
  al.'s~\cite{strahan_self-evacuation_2018} telephone interviews data (n=457).}
  \label{fig:archetypes-baseline-gender-age}
\end{figure*}

Figure~\ref{fig:archetypes-baseline-gender-age} shows two different views of
$\M$, by gender (\ref{fig:archetypes-baseline-gender}), and by age groups
(\ref{fig:archetypes-baseline-age}). A third view of the data by household types
is omitted here for brevity. In subsequent steps, we use the distribution of
archetypes by demographic signature (\texttt{<age,gender,hhtype>}) given by $\M$
here, to probabilistically assign archetypes to individuals of a different
region (to the original group), based on that region's demography.
 \subsection{Synthesising the population for a region}
\label{subsec:population-synthesis}

The aim of population synthesis is to construct virtual instances of persons and
households and assign to them attributes of a real population, such as age, sex,
and relationships, in such a way that the virtual population is a good match,
statistically, to the real population, with respect to those attributes. The
population attributes being matched are typically taken from marginal
distributions available in census tables.

We used a sample-free approach to population synthesis that does not rely on a
disaggregate sample from the original population to seed the algorithm, as is
required by methods like Iterative Proportional Updating
(IPU)~\cite{ye_methodology_2009}. This is particularly useful in the Australian
context where samples are not freely available. Joint marginal distributions for
exactly the desired attributes of the population can readily be downloaded from
the Australian Bureau of Statistics (ABS)
website\footnote{\url{https://abs.gov.au}} using their TableBuilder tool, making
it unnecessary to use techniques like Iterative Proportional Fitting
(IPF)~\cite{deming_least_1940,fienberg_iterative_1970} for deriving such.
The technique we used is described in~\cite{wickramasinghe_heuristic_2018} and
summarised below. Full code for the population synthesis algorithm, along with
instructions for building a synthetic population for any region of Australia are
available on GitHub from \url{https://github.com/agentsoz/synthetic-population}.
A similar approach to ours was proposed in \cite{Huynh2016} for 2006 Australian
census. However, that work is not suitable any more as different household
categorisations are used 2011 onwards. Additionally, that work assumes a family
to be a household. One used here constructs multi-family households, and is more
informative.

Our population synthesiser uses the census data available from the ABS, which
provides population distributions and ESRI Shapefiles of statistical areas. The
Australian Statistical Geography Standard
(ASGS)\footnote{\url{https://www.abs.gov.au/websitedbs/D3310114.nsf/home/Australian+Statistical+Geography+Standard+(ASGS)}}
divides Australia into regions at different granularities. Here, we use data at
Statistical Area 2 (SA2) granularity, which roughly coincides with suburbs and
localities. The person-level data is a joint distribution of the number of
persons under \textit{age}, \textit{gender} and \textit{family/household
relationship} categories. There are eight age categories (\texttt{0--14, 15--24,
25--39, ..., 100+}), two genders (\texttt{Male} and \texttt{Female}) and eight
family/household relationship categories (\texttt{Married}, \texttt{Lone
Parent}, \texttt{Under 15 Children}, \texttt{Student}, \texttt{Over 15
Children}, \texttt{Relatives}, \texttt{Lone Person} and \texttt{Group household
member}). This gives a total of 128 categories ($8 \times 2 \times 8$). The
household-level data is a joint distribution of number of households under eight
\textit{household size} (\texttt{1, 2, ..., 8 or more persons}) and 14
\textit{family household composition} categories, giving a total of 96
categories ($8\times 14$). \texttt{Lone person} \texttt{households} and
\texttt{group} \texttt{households} are the only two non-family categories in the
family household composition categories. The remaining 12 relate to familial
compositions, which represent the number of families (\texttt{1, 2} and
\texttt{3 or more}) in a household and the type of its primary family
(\texttt{Couple with no children}, \texttt{Couple with children}, \texttt{One
parent family} and \texttt{Other family}). An example of a household category is
\texttt{5 persons, 2 Families: Couple with children} that consists of 5 persons in two families where the primary family is a couple that has children.

\newcommand\Tau{\mathcal{T}}
\begin{algorithm} \caption{Population Synthesis Algorithm} \label{alg:popsynth} \SetKwInOut{Input}{Input}\SetKwInOut{Output}{Output}\Input{Persons marginal distribution ($D_p$)\newline Households marginal distribution ($D_h$)\newline SA2 age distribution ($A$)\newline List of SA2s ($S$)\newline Dwellings geographical data ($G$)}\Output{Population($\Tau$)}

  $(D_h, D_p) \leftarrow$ \texttt{CleanData($D_h$, $D_p$)} \\
  $\Tau \leftarrow \emptyset;$\\
  \ForEach{s in S}{
   $\tau \leftarrow \emptyset; d_h \leftarrow D_h(s); d_p \leftarrow D_p(s);$\\
  $P \leftarrow$ \texttt{InstantiatePersons($d_p$)};\\
  $F \leftarrow$ \texttt{FormBasicFamilies($P, d_h$)};\\
  $\tau \leftarrow \tau + $ \texttt{FormLonePersonHouseholds($P, d_h$)};\\
  $\tau \leftarrow \tau + $ \texttt{FormGroupHouseholds($P, d_h$)};\\
$H \leftarrow $ \texttt{FormHhsWithPrimaryFamilies($F, d_h$)} \tcp*{all family households}
$H \leftarrow $ \texttt{AddNonPrimaryFamilies($F,H$)};\\
$H \leftarrow $ \texttt{AddChildren($P,H$)};\\
  $H \leftarrow $ \texttt{AddRelatives($P,H$)};\\
$\tau \leftarrow \tau + H;$\\
  $\tau \leftarrow$ \texttt{AssignAgeToPersons($\tau, A(s)$)};\\
$\tau \leftarrow $\texttt{AssignDewellingsToHouseholds($\tau, G(s)$)};\\
  $\Tau \leftarrow \Tau + \tau$;\\
  }
\end{algorithm}

Our algorithm uses the following assumptions, made by us, and
heuristics derived from the census data descriptions. A couple is assumed to be
a pair of a \texttt{married} \texttt{male} and a \texttt{female} from the same
age category or the \texttt{female} from one age category younger than the
\texttt{male}. Children are at least 15 years younger than the youngest parent.
A parent can have up to eight children. A \texttt{one parent family} consists of
a \texttt{lone parent} and at least one child. A family consisting of two or
more \texttt{relatives}, like two siblings living together, is called an
\texttt{other family}. A family nucleus can also have \texttt{relatives} like a
single older parent or a sibling living together. The primary family of a
multi-family household is decided in the priority order of \texttt{couple with
children} and then \texttt{one parent} families. If there are two families of
the same type, one with more children gets priority. \texttt{Couple with no
children} and \texttt{other family} have the lowest but equal priority.

Algorithm~\ref{alg:popsynth} illustrates the process of synthesising the
population according to the above specifications. The data cleaning checks
whether the persons distribution has the required numbers to generate all the
households and families in the households distribution, and fixes any
discrepancies. As the data does not fully describe secondary and tertiary
families, the cleaning process cannot fix all the discrepancies. The missing
persons required for families are created probabilistically in the later
population synthesis steps. A \textit{basic family structure} is the minimum
configuration of persons required by the family/household type. The age of a
person is determined based on the age category and the distribution of persons
by age within a.

Algorithm~\ref{alg:popsynth} was validated by generating the populations for 306
non-empty SA2s in the 2016 Greater Melbourne area census. As the algorithm uses
the household-level joint distribution as the reference, household-level
distributions of the generated populations match perfectly with the input data.
Thus, the evaluation focused on validating the generated person-level data
distributions against the corresponding input census distributions using the
Freeman-Tukey's goodness of fit test (FT test)~\cite{Freeman1950} and the
Standardised Absolute Error (SAE)~\cite{Voas2001}. Further, all the tests were
performed after removing impossible person-types like \texttt{married under 15
children} from the person-level distributions. The FT test results showed that
the distributions cannot be concluded to be different by rejecting the null
hypothesis for any of the 306 SA2 populations at 0.05 significance level, in
fact, most p-values were well above 0.05, with 291 SA2s having p-values above
0.9. For the SAE test, 298 SA2s gave errors less than 0.05. Among the eight bad
SA2s (out of 306) were four industrial SA2s with very small populations and
significant census data inconsistencies. The remaining four were residential
areas, and their worst SA2 had an SAE of 0.061, which is still well below the
0.1 SAE acceptance threshold proposed in~\cite{Lovelace2015}.

Geographical locations of households are determined by randomly allocating them
to dwellings within an SA2. Vicmap
data\footnote{\url{http://services.land.vic.gov.au/landchannel/content/productCaalogue}}
provides geographical coordinates and addresses of dwellings in Victoria. A
dwelling's SA2 is obtained by finding the SA2-geographical-polygon containing
it.

The output of this step was a set of \texttt{.csv} files that constitute a
relational database of unique persons, in unique families, in unique households,
in unique dwellings. Importantly, each person in the synthetic population has
its own set of attributes including \texttt{<age,gender,hhtype>}
\footnote{Note that the \texttt{hhtype} household type names in Strahan et al.'s
data~\cite{strahan_self-evacuation_2018} do not directly match the names used by
the population synthesis algorithm~\cite{wickramasinghe_heuristic_2018}, however
this was easily resolved by mapping one to the other since conceptually they
deal in the same groupings.}
among others, which we use to perform the demographic match to and
assignment of archetypes, using the matrix
$\M$ from Section~\ref{subsec:archetypes-probabilities}, as described next.
 \subsection{Assigning archetypes to the population}\label{subsec:archetypes-assignment}

The function for probabilistically assigning archetypes to the synthetic
population of a new region is given in
Algorithm~\ref{alg:archetypes-assignment}. The function operates one by one on
the set of individuals $\I$ in the input synthetic population (lines 1--9). For
a given individual $i$, it first extracts its demographic signature of the form
\texttt{<age,gender,hhtype>} (line 2), and the associated archetypes
probabilities $p$ for that signature from matrix $\M$ (line 3). The obtained
vector has size $\card{p}=8$, being the probabilities of the seven archetypes
plus the extra \UT type. Vector $p$ is then normalised to $1$ (line 4), and
re-assigned the cumulative sum of its elements (line 5). The function then picks
a random number $r$ in the range [0,1] (line 6), and finds the vector index $a$
for which the cumulative probability exceeds $r$ (line 7). Finally the archetype
associated with index $a$ is assigned to individual $i$ (line 8).

\begin{algorithm}[h]
\SetKwData{a}{a}
\SetKwData{i}{i}
\SetKwData{s}{s}
\SetKwData{p}{p}
\SetKwData{r}{r}
\SetKwData{TRUE}{TRUE}
\SetKwData{arch}{a.archetype}
\SetKwFunction{Colnames}{Colnames}
\SetKwFunction{GetSignature}{GetSignature}
\SetKwFunction{GetSignatureRow}{GetSignatureRow}
\SetKwFunction{Sum}{Sum}
\SetKwFunction{CumSum}{CumSum}
\SetKwFunction{Runif}{Runif}
\SetKwFunction{Match}{Match}
\SetKwInOut{Input}{Input}\SetKwInOut{Output}{Output}
\Input{Set of individuals $\I$, probabilities matrix $\M$}
\Output{Set of assigned archetypes $\A$}
\BlankLine
\For{$i\leftarrow 1$ \KwTo $\card{\I}$}{
    \s$\leftarrow$\GetSignature{$\I\left[\i\right]$}\tcp*{<age,gender,hhtype>}
    \p$\leftarrow$\GetSignatureRow{\s, $\M$}\tcp*{get probabilities}
    \p$\leftarrow$\p$/$\Sum{\p}\tcp*{normalise probabilities}
    \p$\leftarrow$\CumSum{\p}\tcp*{cumulative sum of probs}
    \r$\leftarrow$\Runif{1}\tcp*{random number in [0,1]}
    \a$\leftarrow$\Match{\TRUE, \p $>$ \r}\tcp*{select index}
    $\A\left[\i\right]\leftarrow$\Colnames{$\M$}[a]\tcp*{assign archetype}
  }
\caption{Algorithm to probabilistically assign archetypes to individuals of a synthetic population}
\label{alg:archetypes-assignment}
\end{algorithm}
 \subsection{Assigning attitudes to archetypes}
\label{subsec:archetypes-attitudes}

The matrix $\B$ in Table~\ref{tab:archetypes-attitudes} shows the different
bushfire-related attitudes (in the range $[0,1]$) that are assigned to
individuals of the different archetypes. Each row captures the relative
difference in the value that each archetype places on the attitude of the first
column, irrespective of all other considerations. Each column captures how a
given archetype values the different attitudes relative to each other. Higher
values indicate a higher level of importance placed on an attitude.
For instance, a \CG person  (column 2) places a lot more importance on the
sight of fire (\VisibleFire=$0.5$) than smoke (\VisibleSmoke=$0.2$). Also this
kind of person is impacted more by some official messages received from the
emergency services (\EmergencyWarning=$0.33$, \EvacuateNow=$0.3$), than the
sight of smoke.
On the other hand, an official \EmergencyWarning (row 7) is considered important
by \CE, \CG, and \WW archetypes, but is of no consequence to the other types
(\EmergencyWarning=$0.0$).

For each archetype, each attitude-value pair in the respective column of matrix
$\B$ was assigned to every individual of that type in the population. The only
values from matrix $\B$ that varied between individuals of the same archetype
were \ThresholdInitial and \ThresholdFinal which represent the risk-averseness
of an individual. For these attitudes, the values in the matrix were taken as
the mean $\mu$ of a normal distribution with standard deviation $\sigma=0.1$,
and the values assigned to the individuals were samples drawn from this
distribution. The standard deviation $\sigma$ is not calibrated here, and the
value was chosen manually to give sufficient variation in attitudes across
archetypes.

\begin{table}[!t]
  \centering
  \caption{Archetypes behaviour attitudes matrix $\B$. The column names stand
  for CE:~\CE, CG:~\CG, TD:~\TD, WW:~\WW, RD:~\RD,DE~\DE, and EI:~\EI.}
  \label{tab:archetypes-attitudes}
  \npdecimalsign{.}
  \nprounddigits{3}
  \begin{tabular}{
    |
    p{0.25\columnwidth}|
    n{1}{3}@{\hspace{10pt}}
    n{1}{3}@{\hspace{10pt}}
    n{1}{3}@{\hspace{10pt}}
    n{1}{3}@{\hspace{10pt}}
    n{1}{3}@{\hspace{10pt}}
    n{1}{3}@{\hspace{10pt}}
    n{1}{3}@{\hspace{10pt}}
    |
  }
  \hline
  {\bf Attitude} & {\bf CE} & {\bf CG} & {\bf TD} & {\bf WW} & {\bf RD} & {\bf DE} & {\bf EI}\\
  \hline
  \hline
\VisibleSmoke & 0.3 & 0.2 & 0.0 & 0.1 & 0.0 & 0.2 & 0.0 \\
\VisibleEmbers & 0.5 & 0.4 & 0.2 & 0.4 & 0.2 & 0.4 & 0.3 \\
  \VisibleFire & 0.6 & 0.5 & 0.5 & 0.5 & 0.3 & 0.5 & 0.4 \\
\Advice & 0.103 & 0.091 & 0.0 & 0.088 & 0.0 & 0.0 & 0.\\
  \WatchAndAct & 0.176 & 0.172 & 0.0 & 0.167 & 0.0 & 0.0 & 0.\\
  \EmergencyWarning & 0.337 & 0.326 & 0.0 & 0.313 & 0.0 & 0.0 & 0.\\
  \EvacuateNow & 0.302 & 0.295 & 0 & 0.285 & 0.356 & 0 & 0.437 \\
\hline
  \ThresholdInitial & 0.3 & 0.3 & 0.6 & 0.3 & 0.4 & 0.3 & 0.4 \\
  \ThresholdFinal & 0.3 & 0.5 & 0.8 & 0.7 & 0.5 & 0.3 & 0.9 \\
  \hline
  \end{tabular}
\end{table}

The choice of attitudes to include in $\B$ was influenced by the functionality
afforded by the underlying simulation system. For instance, while it is known
that the sight and sounds of emergency services vehicles in the vicinity impacts
the anxiety levels of people in different ways~\cite{mclennan_wait_2012},
this attitude was not included
in $\B$ because movements of emergency vehicles are not included in the
underlying simulation model presented here. It is expected that as the model is
refined in future work and the functionality of the system expanded, other
missing but significant attitudes will be added to the model.

The attitudes' values in $\B$ were initially chosen manually based on the
derived understanding of attitudes of the archetypes from Strahan et al.'s
paper~\cite{strahan_self-evacuation_2018}. These were subsequently calibrated to
match more generally the levels of responses typically seen in bushfires in
Australia, as described in Section~\ref{subsec:attitudes-calibration}.
 \subsection{Calibrating attitudes of archetypes}
\label{subsec:attitudes-calibration}

The purpose of the calibration effort was to tune the model such that the
observed reactions to environmental stimuli matched expected levels of responses
to such stimuli from previous bushfires.
We will use the example of calibrating the responses to official emergency
messages in this section, but note that the same method applies also for other
stimuli such as the sight of smoke.
Table~\ref{tab:archetypes-attitudes-uncalibrated} shows a matrix $\V$ which is a
subset of rows of matrix $\B$ related to the attitudes towards official
emergency messages.

\begin{table}[!t]
  \centering
  \caption{Matrix $\V$ corresponding to select rows of behaviour attitudes
  matrix $\B$ with initial uncalibrated values. The column names stand for
  CE:~\CE, CG:~\CG, TD:~\TD, WW:~\WW, RD:~\RD,DE~\DE, and EI:~\EI.}
  \label{tab:archetypes-attitudes-uncalibrated}
  \npdecimalsign{.}
  \nprounddigits{2}
  \begin{tabular}{
    |
    p{0.25\columnwidth}|
    n{1}{3}@{\hspace{10pt}}
    n{1}{3}@{\hspace{10pt}}
    n{1}{3}@{\hspace{10pt}}
    n{1}{3}@{\hspace{10pt}}
    n{1}{3}@{\hspace{10pt}}
    n{1}{3}@{\hspace{10pt}}
    n{1}{3}@{\hspace{10pt}}
    |
  }
  \hline
{\bf Attitude} & {\bf CE} & {\bf CG} & {\bf TD} & {\bf WW} & {\bf RD} & {\bf DE} & {\bf EI}\\
  \hline
  \hline
  \Advice           & 0.25 & 0.20 & 0.00 & 0.20 & 0.00 & 0.25 & 0.00\\
  \WatchAndAct      & 0.25 & 0.20 & 0.00 & 0.20 & 0.00 & 0.25 & 0.00\\
  \EmergencyWarning & 0.30 & 0.30 & 0.00 & 0.30 & 0.00 & 0.30 & 0.00\\
  \EvacuateNow      & 0.40 & 0.40 & 0.00 & 0.35 & 0.20 & 0.50 & 0.10\\
  \hline
  \end{tabular}
\end{table}

The values in $\V$ are the uncalibrated values that were
manually assigned to these rows initially from derived understanding of
attitudes of the archetypes from Strahan et al.’s paper [14].
We then applied the calibration process described in
Algorithm~\ref{alg:attitudes-calibration} to $\V$, to get the final calibrated
rows of $\B$ reported in Table~\ref{tab:archetypes-attitudes}.
Algorithm~\ref{alg:attitudes-calibration} takes these best-guess values and
adjusts them, such that the initial response rates achieved across the
population for the different types of emergency messages match generally
observed response rates $\R$ in previous bushfires being $\Advice=1\%$,
$\WatchAndAct=5\%$, $\EmergencyWarning=30\%$, and $\EvacuateNow=40\%$
\cite{mclennan_should_2019}.

$\R$ is a matrix of the same dimensions as $\V$ and is obtained by matrix
multiplication of the row vector of probabilities $\begin{Bmatrix}0.01 & 0.05 &
0.3 & 0.4\end{Bmatrix}$ with the column vector of archetypes $\begin{Bmatrix}1 &
1 & 1 & 1 & 1 & 1 & 1\end{Bmatrix}$. The vector $\D$ is the distribution
of archetypes in the synthetic population.

\begin{algorithm}[h]
\SetKwData{vd}{$U$}
\SetKwData{i}{i}
\SetKwData{s}{s}
\SetKwData{p}{p}
\SetKwData{r}{r}
\SetKwFunction{Nrow}{Nrow}
\SetKwFunction{Sum}{Sum}
\SetKwFunction{QNorm}{QNorm}
\SetKwInOut{Input}{Input}\SetKwInOut{Output}{Output}
\Input{Matrix of uncalibrated values $\V$, Archetypes distribution vector $\D$,
Required response rates matrix $\R$, Vector of initial threshold means $\Tau$,
Standard deviation of initial threshold values $\sigma$}
\Output{Matrix $U$ of calibrated values}
\BlankLine
\vd$\leftarrow\V$;\\
\For{$\r \leftarrow 1$ \KwTo \Nrow{$U$}}{
  $\vd[r,]\leftarrow\vd[r,]\cdot\D$\tcp*{weighted by archetypes distribution}
  $\vd[r,]\leftarrow\vd[r,]/$\Sum{$\vd[r,]$} \tcp*{normalise row}
}
$\vd\leftarrow\vd\cdot\R$ \tcp*{distribution adjusted response rates}
\For{$\r \leftarrow 1$ \KwTo \Nrow{$U$}}{
  $\vd[r,]\leftarrow$\QNorm{$\vd[r,],\Tau,\sigma$}\tcp*{P(X < A)}
}
\caption{Algorithm to calibrate matrix $\V$ to produce the values used in matrix
$\B$.}
\label{alg:attitudes-calibration}
\end{algorithm}

Algorithm~\ref{alg:attitudes-calibration} proceeds by first calculating the
normalised joint distribution by row, i.e, for each emergency message type,
given by $\V\cdot\D$ (lines 1-5). This distribution is then multiplied by $\R$
to effectively get the desired response rates by emergency message, by
archetypes' distribution, by initial estimate $\V$ (line 6). Since the initial
response thresholds of archetypes are normally distributed about the means
$\Tau$ and standard deviation $\sigma$ (see
Section~\ref{subsec:archetypes-attitudes}), we can therefore use the obtained
probability $U[r,c]$ for a given message type (row $r$) and archetype (column
$c$), to calculate the value for the normal distribution $\mu=\Tau[c],\sigma$ at
which the cumulative distribution matches the probability $U[r,c]$ (line 8).
\footnote{The \texttt{qnorm} function in the R statistical package does
precisely this and was used here.}
The output of the process is matrix $U$ which constitutes the relevant rows in
$\B$ (see Table~\ref{tab:archetypes-attitudes}).

\begin{table}[!t]
\centering
\caption{Actual response rates achieved in the simulation after calibrating
using Algorithm~\ref{alg:attitudes-calibration}. Numbers reflect the percentage
of all persons who received the message and responded. Columns do not add up to
100.}
\label{tab:response-rates}
\begin{tabular}{|l|p{0.2\textwidth}|p{0.2\textwidth}|p{0.2\textwidth}|}
\hline
{\bf Message} & {\bf Response (\%) desired} &
{\bf Response (\%) pre-calibration} &
{\bf Response (\%) post-calibration}\\
\hline
\hline
\Advice           & 1.0  & 16.3 & 1.0 \\
\WatchAndAct      & 5.0  & 16.3 & 5.1 \\
\EmergencyWarning & 30.0 & 31.8 & 31.5 \\
\EvacuateNow      & 40.0 & 45.0 & 42.1 \\
\hline
\end{tabular}
\end{table}

To verify that the desired response rates are achieved after calibration, we ran
the simulation several times without any bushfire, but instead each time sending
just one of the four emergency messages to the full population, and recording
the total number of agents whose initial threshold \ThresholdInitial was
breached. The response in this case could therefore be fully attributed to the
individual message that was sent. The actual response rates observed are
reported in Table~\ref{tab:response-rates}.
 \subsection{Assigning activities to the population}
\label{subsec:archetypes-activities}

In this step we address the question: where is the population and what is it doing at the time of the bushfire emergency?
A common approach for capturing regular activities and trips of a population is
activity-based travel demand
modelling~\cite{castiglione_activity-based_2015,horni_multi-agent_2016}. The
typical process here is to use available data about the population on origins
and destinations, timings of trips and activities, and modes of travel, to
estimate the travel demand for a given day. The model is then calibrated against
expected travel times and distances, mode share, and traffic volumes.

For Victoria, the best source of information on population activities is
Victorian Integrated Survey of Travel and Activity
(VISTA)~\footnote{\url{https://transport.vic.gov.au/data-and-research/vista}},
which is a regular survey of 1\% of the population. Together with Australian
census data~\footnote{\url{https://www.abs.gov.au/census}}, VISTA travel dairies
can be successfully used to construct a validated activity-based model, as
demonstrated by the Melbourne Activity-Based Model
(MABM)~\cite{infrastrucure_victoria_model_2018}.
However, VISTA data coverage for regional Victoria--where bushfire risk is
highest--is limited, and models like MABM therefore have limited representation
of traffic patterns in these regions. In any case, census and VISTA data may not
sufficiently capture the population that may be present in the region on the
bushfire day. For instance, townships along the famous Great Ocean Road in
Victoria may see the daytime population surge by 10-fold due to visiting
tourists.

Previous work with emergency services in regional Victoria has looked at an
alternative method that elicits best-guess information from experts about the
makeup of the population and the distribution of their activities on the day, in
the form of an easily specified table, and uses this information to
automatically construct a synthetic population that on the whole matches this
distribution~\cite{robertson_modelling_2019}.
In this model, a plan is generated for each individual based on the population subgroup they belong to.
A set of possible activities that people are typically engaged with in the region is used to define both activity distributions and a set of feasible activity location co-ordinates for each subgroup.
Candidate activities are assigned first with a per time-step procedure that uses start-time probabilities derived from the activity distribution.
The output of this process can be interpreted as a Boolean matrix $B$ with each row representing an activity and each column representing a time-step:

$$
\setcounter{MaxMatrixCols}{20}
B=\begin{pmatrix}
   1 & 0 & 1 & 0 & 0 & 0 & 1 & 0 & 0& 0 & 0 & 1 &\\
   0 & 0 & 0 & 1 & 0 & 1 & 0 & 0 & 0& 0 & 0 & 0 &\\
   0 & 1 & 0 & 0 & 0 & 0 & 0 & 1 & 1& 0 & 0 & 0 &\\
   0 & 0 & 0 & 0 & 1 & 0 & 0 & 0 & 0& 1 & 0 & 0 &\\
   0 & 0 & 0 & 0 & 0 & 0 & 0 & 0 & 0& 0 & 1 & 0 &
  \end{pmatrix}
  $$

A workable ordered plan is then established according to durations associated with each activity (these may also vary by subgroup).
Iteratively, the matrix $B$ is organised according to the length of time remaining from a previously selected activity.
The resultant matrix will then give a feasible day plan:

$$
\setcounter{MaxMatrixCols}{20}
B=\begin{pmatrix}
   1 & 1 & 1 & 1 & 0 & 0 & 0 & 0 & 0& 0 & 0 & 1 &\\
   0 & 0 & 0 & 0 & 0 & 1 & 1 & 1 & 0& 0 & 0 & 0 &\\
   0 & 0 & 0 & 0 & 0 & 0 & 0 & 0 & 1& 1 & 0 & 0 &\\
   0 & 0 & 0 & 0 & 1 & 0 & 0 & 0 & 0& 0 & 0 & 0 &\\
   0 & 0 & 0 & 0 & 0 & 0 & 0 & 0 & 0& 0 & 1 & 0 &
  \end{pmatrix}
  $$
where the unique 1 value in each column represents the planned activity for that time-step.
Start times are then added to each activity by selecting a random time within the corresponding time-step, which across the population avoids waves of trips occurring periodically at each time-step interval.

Once an ordered schedule has been organised, it remains to assign locations to each activity.
Individuals are given a \texttt{home} location, according to their subgroup, which denotes where their plan will start (and end).
The location of each subsequent activity is then iteratively chosen at the locality level, which is essentially a group of addresses in a common area.
The probability of a locality being chosen for the next activity is inversely proportional to its centroid distance from the current activity's locality and proportional to a sum of an allocation weighting assigned to each eligible location within the locality.
Once the locality is chosen, an eligible location within it is chosen at random, and that location's allocation weighting is reduced by one.
Locality-based selection allows some representation of personal preference in location choice, and also reduces the computational requirements of the algorithm as the number of prescribed location options increases.
With correctly selected allocation weights, location selection follows the gravity model of transportation \cite{anderson_gravity_2011} and allows for specific scenarios to be tested in the model, such as an event at a beach resulting in that beach being more popular than others on that particular day.
\footnote{
A population was constructed for Surf Coast Shire using the algorithm from \cite{robertson_modelling_2019} in consultation with domain experts from the Surf Coast Shire Council (SCSC) and the Department of Environment, Land, Water and Planning (DELWP).

Subgroups were defined as \texttt{Resident}, \texttt{ResidentPartTime},
\texttt{VisitorRegular}, \texttt{VisitorOvernight} and \texttt{VisitorDaytime}.
The set of possible activities were \texttt{home}, \texttt{work}, \texttt{beach}, \texttt{shops} and \texttt{other}; the \texttt{Visitor*} subgroups were not eligible for the \texttt{work} activity and thus it did not appear in their activity distribution tables.
The \texttt{other} activity was useful to distinguish between locations that different subgroups would be likely to attend.
An \texttt{other} activity for a \texttt{Resident} may be at a school, whereas for \texttt{Visitor*} individuals it would likely map to local landmarks or national parks.
In the output plans, for the two \texttt{Resident*} subgroups the cumulative
effect of \texttt{work} having a longer duration than other activities resulted
in $\approx5\%$ discrepancy towards the end of the peak working period, when
compared against the desired input distribution. In the other subgroups where
\texttt{work} was not a valid activity, the generated plans matched the input
distributions to a maximum error of $\approx 1\%$, with the error spread evenly
throughout the `busy' part of the day.

Per-address data was supplied by SCSC and allowed the \texttt{home} locations for the \texttt{Resident*} population to be distributed accurately throughout the region.
\texttt{VisitorRegular} and \texttt{VisitorOvernight} individuals had their \texttt{home} mapped to hotels, camp grounds and some allocated residential addresses to reflect the tendency for visitors to stay in lodging or home stay type arrangements.
\texttt{VisitorDaytime} had a small number of source nodes set on the edges of the region, with the majority allocated to the eastern side (traffic inbound from Melbourne).
An additional parameter in the algorithm allows for the likelihood of moving between localities for new activities to be set at a per-subgroup level.
In this case, the \texttt{Visitor*} individuals were more likely to move across the region throughout the day.
}
 
This method allows rich activity-based populations to be constructed through
the relatively simple specification of the input matrix of daily activities,
however it relies heavily on expert opinion and therefore results can vary
significantly depending on the experts consulted.

In the end what is required is very likely a scheme that takes the best of the
above two approaches, by generalising trip patterns found in urban and peri-urban
samples of VISTA data to regional areas, and using expert opinion as well as
regional traffic data counts to calibrate the result. This is an open research
area and no complete solution of this kind is available at present to the best
of our knowledge.

For the purpose of this work, and given that activity modelling is not the focus
here, we will simplify the issue and assume that the synthetic individuals are
all located in their homes at the time of the bushfire. This is rather
unrealistic, but captures the expected clustering of vehicles in densely
populated areas. The assignment of individuals to street addresses, while a
simplification, is nevertheless non-trivial and is performed by the Address
Mapper
algorithm~\footnote{\url{https://github.com/agentsoz/synthetic-population}}
during the population synthesis step described in
Section~\ref{subsec:population-synthesis}.
Overall, while we do not assign activities with locations, each individual in
the population is still assigned a set of coordinates that are used during
evacuation to drive to the location of dependants, and/or to evacuation points,
as described in Table~\ref{tab:person-coordinates}.

\begin{table}[!t]
\centering
\caption{Coordinates that are assigned to individuals in the synthetic population.}
\label{tab:person-coordinates}
\begin{tabular}{|l|p{0.6\textwidth}|}
\hline
{\bf Coordinates} & {\bf Description}\\
\hline\hline
\texttt{Geographical.Coordinate} &
Home coordinates assigned by the population synthesis algorithm of
Section~\ref{subsec:population-synthesis} to be statistically correct with
respect to the distribution of households in SA1 areas given by the 2016 census.
\\
\texttt{EvacLocationPreference} &
An individual's preferred out-of-region evacuation destination, when no
evacuation destination is suggested by the emergency services messaging;
typically assigned by drawing with replacement from a small set of coordinates
of nearby townships.
\\
\texttt{InvacLocationPreference} &
An individual's preferred within-region evacuation destination when evacuation
to \texttt{EvacLocationPreference} is not possible, such as due to heavy
congestion and/or a road blockages; typically assigned to the nearest town
centre from \texttt{Geographical.Coordinate}.
\\
\texttt{HasDependentsAtLocation} &
Location of dependants, assigned only to those individuals that have dependants;
coordinates are assigned by randomly drawing without replacement from the set of
home coordinates of \DE individuals, and if exhausted from a circle of radius
5km around \texttt{Geographical.Coordinate}.
\\
\hline
\end{tabular}
\end{table}
 \subsection{Assigning behaviours to the population}
\label{subsec:bdi-behaviours}

The question we address in this section is: what will the population do when the
bushfire emergency is upon it? The underlying paradigm we use is called the
Belief-Desire-Intention (BDI) model of agency, based in folk psychology and
initial work by Bratman on rational systems~\cite{bratman_intention_1987} and
Dennett on intentional stance~\cite{dennett_intentional_1989}. This work has
since been translated to a computational model of bounded
rationality~\cite{cohen_intention_1990,rao_modeling_1991,rao_bdi_1995} and
many systems exist today in the BDI-tradition such as
Jack~\cite{winikoff_jack_2005}, Jason~\cite{bordini_programming_2007}, and
Jadex~\cite{pokahr_programming_2014}--and one we use in this work called
Jill
\footnote{Our motivation for using Jill is that it is a fast and lightweight
open-source BDI engine designed for large-scale simulations with millions
of agents. Jill is available at \url{https://github.com/agentsoz/jill}.}.

The BDI premise is that people act according to their \textit{beliefs}--factual
or not--about the world. We hold \textit{desires}--often conflicting--about
states of affairs we would like to bring about, and these desires determine our
goals. Our \textit{intentions} are goals we are committed to achieving. And
while we do form \textit{plans} of action to achieve goals, these are usually
fairly high level and change as we go, adapting to unforeseen situations.

The BDI computational model is a program of the type
$\epsilon:\tau\leftarrow\phi$ that provides a plan of action $\phi$ for handling
goal $\epsilon$ when the \textit{context} condition $\tau$ holds. The plan
$\phi$ is an abstract recipe written by a programmer and prescribes a course of
action for achieving the goal $\epsilon$. Several such plans might be supplied
by the programmer, for achieving the same goal $\epsilon$ but in different
situations, captured by different context conditions. The plan $\phi$ could be
composed of a sequence of complex tasks, or sub-goals, that may invoke other
sub-plans to achieve them, in a hierarchical fashion.

\begin{figure}[!t]
  \centering
  \leavevmode\beginpgfgraphicnamed{archetypes-gptree-ext}
\begin{tikzpicture}[level distance=10mm]
\tikzstyle{textstyle}=[font=\ttfamily\fontsize{8}{0}\selectfont]
\tikzstyle{plan}=[textstyle,rounded corners=1mm,minimum height=2em,draw]
\tikzstyle{goal}=[textstyle,draw,ellipse,inner sep=0mm,minimum height=2em]
\tikzstyle{level 2}=[level distance=8mm,sibling distance=40mm]
\tikzstyle{level 3}=[level distance=10mm,sibling distance=17mm]
\tikzstyle{level 4}=[level distance=12mm,sibling distance=10.5mm]

\node[goal] {Response}
	child {node[plan] {FullResponse}
			child {node[goal] {InitialResp}
			child{node[plan] {DepsNear}
				child{node[goal] {Go(d)}}
				child[densely dashed]{node[goal] {Go(h)}}
				child[missing] {node {}}
			}
			child{node[plan] {DepsFar}
				child[missing] {node {}}
				child[missing] {node {}}
				child{node[goal] {Go(h)}}
				child{node[goal] {Go(d)}}
				child[densely dashed]{node[goal] {Go(h)}}
				child[missing] {node {}}
			}
			child{node[plan] {NoDeps}
				child[missing] {node {}}
				child[missing] {node {}}
				child[densely dashed]{node[goal] {Go(h)}}
			}
			}
			child {node[goal] {FinalResp}
				child{node[plan] {Defend}}
				child{node[plan] {Leave}
					child{node[goal] {Go(e|*)}}
					child[missing] {node {}}
				}
			}
	}
;
\end{tikzpicture}
\endpgfgraphicnamed
   \caption{Archetypes' response behaviour BDI goal-plan tree. A goal (ellipse)
  is achieved by any one of its children plans.  Each plan (box) is a sequence
  of goal actions from left to right. \texttt{Go(?)} is a parameterised goal
  (achieved by plan \texttt{Goto} not shown for brevity) that takes a
  destination to go to (\texttt{h} is home location, \texttt{d} is dependants
  location, and \texttt{e} is evacuation location). Dashed goals are optionally
  executed with some probability.}
  \label{fig:archetypes-gptree}
\end{figure}
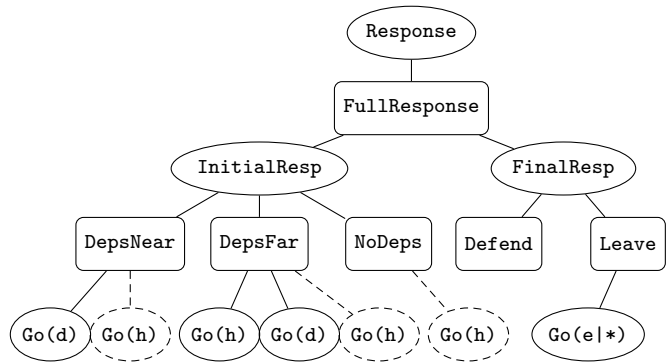

Figure~\ref{fig:archetypes-gptree} shows the BDI goal-plan tree that we assign to each archetype person.
Note that this behaviour is applied to every individual in the population
irrespective of its archetype. This is because while Strahan et al.'s
archetypes~\cite{strahan_self-evacuation_2018} are useful for understanding
behaviour leading up to the decision to leave, they tell us little about what
people will do once on the road. For this we rely on expert knowledge and
accounts from reports and research papers. The goals and plans shown in
Figure~\ref{fig:archetypes-gptree} are the final result of an iterative process
of co-development with experts.
The goal-plan tree captures sufficient variability in the behaviour of the
population in a compact representation as shown in
Table~\ref{tab:behaviour-examples}.

\begin{table}[!t]
  \centering
  \caption{Examples of behaviour afforded by the BDI goal-plan tree of
  Figure~\ref{fig:archetypes-gptree} once a person has decided to act on the
  situation. Behaviour observed in the simulation has less variation since all
  persons are assumed to be home at the time of the bushfire event (see
  Section~\ref{subsec:archetypes-activities}).}
  \label{tab:behaviour-examples}
  \begin{tabular}{|l|p{0.4\columnwidth}|p{0.4\columnwidth}|}
    \hline
    {\bf Id} & {\bf Initial response} & {\bf Final response}\\
    \hline\hline
    \texttt{A1} & Has no dependants and goes home to prepare &
    Leaves from home\\\hline
    \texttt{A2} & Goes home first, then collects dependants, then goes back
     home to wait & Leaves from home with dependants\\\hline
    \texttt{A3} & Attends to dependants and stays there & Leaves from
     dependants location\\\hline
    \texttt{A4} & Collects dependants then goes home & Leaves from home with
     dependants\\\hline
    \texttt{A5} & Has no dependants and continues with activity & Stops
     activity and leaves\\\hline
    \texttt{A6} & Goes home first, then collects dependants, and waits there &
     Leaves from dependants location \\\hline
  \end{tabular}
\end{table}

The goal-plan tree contains three high level goals \texttt{InitialResp},
\texttt{FinalResp}, and \texttt{Response} that are triggered depending on
whether the internal anxiety barometer of the individual crosses the initial
threshold \texttt{ThresholdInitial}, or final threshold \texttt{ThresholdFinal},
or both together, given observations and their values (see
Table~\ref{tab:archetypes-attitudes}).

The way the values of external stimuli are combined to calculate an individual's
internal anxiety barometer is as follows. Over the course of the simulation, as
an agent becomes aware of the unfolding bushfire situation, only the highest
value of any environmental observation (\texttt{Visible*} attitudes in
Table~\ref{tab:archetypes-attitudes}) is kept. Similarly for emergency messages
(remaining rows of Table~\ref{tab:archetypes-attitudes}), only the highest value
of any received messages is maintained. The individual's anxiety barometer value
is then given by the sum of these two values at any given time.

During the initial response, people who have dependants will always go and
attend to them, but the behaviour is slightly different depending on whether
their current location is closer to the dependants (plan \texttt{DepsNear}) or
home (plan \texttt{DepsFar}). If closer to  dependants, they always attend to
them, and then optionally go home afterwards. If closer to home, they go home
first (goal \texttt{Go(h)}) then to dependants (goal \texttt{Go(d)}), and then
again optionally back home. If on the other hand they do not have dependants
(plan \texttt{NoDeps}), they either go home or continue doing what they are
doing.

The final response is always to either stay and defend the residence, or leave.
In the model, those with dependants will always leave, but some others with
higher thresholds of response--such as {\EI} types--are more likely to stay. The
leaving behaviour realised by the $\texttt{Leave}$ plan is in itself
non-trivial, since things can go wrong along the way, such as if the person
leaving gets stuck in traffic congestion or comes across an accident or road
closure (all possibilities in the simulation model). In such cases, the
evacuating agent will try re-routing to find an alternative path to the
evacuation destination $\texttt{e}$ up to three times
\footnote{Anecdotal accounts by some field researchers who conducted
post-bushfire interviews with residents who had survived major bushfires in
Australia over the period 2009-2014~\cite{mclennan_at-risk_2015} suggested that
residents who had self-evacuated in vehicles tended to make up to three thwarted
attempts to drive to their initially chosen destination before abandoning that
goal and deciding to drive to another presumed place of safety.},
before giving up, and trying for an alternative ``in-vac''
location $\texttt{i}$ within the region that it might consider safe. If that
also fails for similar reasons, then it tries to go to its home location
$\texttt{h}$, failing which it stops, and is counted in the list of stranded
persons in the final statistics calculated from the simulation outputs.
 \subsection{Putting it all together}
\label{subsec:ees}

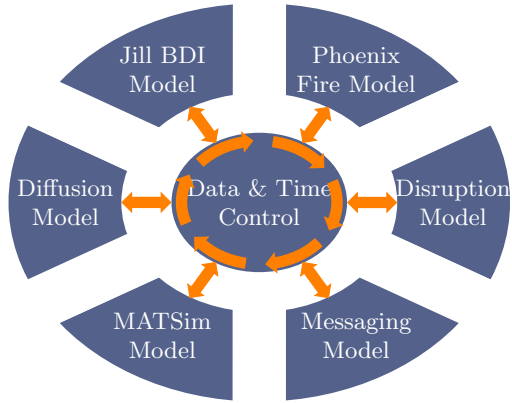
\begin{figure}[!t]
\centering
\leavevmode\beginpgfgraphicnamed{ees-components-ext}
\def\models#1#2#3#4#5{
\resetcolorseries[#2]{#3}
  \edef\startangle{90}
  \foreach [
    remember=\endangle as \startangle,
    evaluate=\i as \endangle using {\startangle-(\csname#1\endcsname[\i-1]/100*360)},
    evaluate=\halfangle using {(\endangle-\startangle)/2+\startangle},
  ] \i in {1,...,#2} {\fill[{#3!![\i]}] (0,0) --++(\startangle:#4) arc (\startangle:\endangle:#4);
\node[white,text width=1.8cm,align=center] at (\halfangle:#4-0.75cm) {\pgfmathparse{\csname#5\endcsname[\i-1]}\pgfmathresult};
\draw[white, line width=7mm](0,0)--++(\startangle:#4+2pt);
    \draw[white, line width=7mm](0,0)--++(\endangle:#4+2pt);
  }

  \fill[white](0,0)circle [radius=#4*0.55];\fill[darkblue!70](0,0)circle [radius=#4*0.35];\node[white,text width=2cm,align=center] at (0,0) {Data \& Time Control};

  \foreach [
    remember=\endangle as \startangle,
    evaluate=\i as \endangle using {\startangle-(\csname#1\endcsname[\i-1]/100*360)},
    evaluate=\halfangle using {(\endangle-\startangle)/2+\startangle},
  ] \i in {1,...,#2} {\draw[orange,line width=1.5mm,preaction={orange,-triangle 90,thin,draw,shorten >=-1mm}] ++(\startangle-10:#4*0.31) arc[radius=#4*0.31, start angle=(\startangle-10), end angle=(\endangle+10)];
\draw[orange,line width=1.5mm,preaction={orange,thin,draw,<->,>=triangle 90, shorten >=-1mm, shorten <=-1mm}] ++(\halfangle:#4*0.38) -- ++(\halfangle:#4*0.14);}
  }

\begin{tikzpicture}[yscale=0.8]
\definecolor{darkblue}{rgb}{0.04,0.12,0.35}
\definecolorseries{colorser}{rgb}{last}{darkblue!70}{darkblue!70}
\newcommand\shares{{16.67,16.67,16.67,16.67,16.67,16.67}}
\newcommand\labels{{
  "Phoenix Fire Model",
  "Disruption Model",
  "Messaging Model",
  "MATSim Model",
  "Diffusion Model",
  "Jill BDI Model"}}
{\small
  \models{shares}{6}{colorser}{3.3cm}{labels}
}
\end{tikzpicture}
\endpgfgraphicnamed
 \caption{Component view of the Emergency Evacuation Simulator.  }
  \label{fig:ees-components}
\end{figure}

The final simulation model is realised in the Emergency Evacuation Simulator as
shown in Figure~\ref{fig:ees-components} and has the following components.

\texttt{MATSim Model}: Contains the
MATSim~\cite{horni_making_2016}~\footnote{\url{https://github.com/matsim-org/matsim}}
traffic model including the road network derived from
OpenStreetMap~\footnote{\url{https://www.openstreetmap.org}}, and the driving
agents representing the population of the region in consideration.

\texttt{Jill BDI Model}: Contains the ``brains'' of the archetypes agents
implemented as BDI agents in the Jill
system~\footnote{\url{https://github.com/agentsoz/jill}} as per the description
in Section~\ref{subsec:bdi-behaviours}. The ``bodies'' of these agents are the
driving agents within the \texttt{MATSim Model}, with the two coupled together
using the BDI-ABM framework~\cite{singh_integrating_2016}.

\texttt{Phoenix Fire Model}: Reads the outputs of the Phoenix RapidFire
simulator (available as a time-varying polygon shape saved in \texttt{geojson}
format), and injects these at the appropriate times into the simulation. The
shapes are received by the \texttt{MATSim Model} that then dynamically adjusts
the penalty on the affected road links such that vehicles subsequently routing
through the network will tend to avoid the fire-affected area.

\texttt{Disruption Model}: Injects user-specified road disruptions--such as road
closures that restrict access, or accidents that constrict traffic flow--at
specified times into the simulation. Information includes the start and end time
of the disruption, the affected road network links, and the impact in terms of
percentage reduction in traffic flow. The information is received by the
\texttt{MATSim Model} that dynamically adjusts the network accordingly so that
vehicles subsequently routing will notice the change.

\texttt{Diffusion Model}: Responsible for modelling the diffusion of
information--such as about the location of the fire or road blockages--through a
social network using standard information diffusion schemes (not covered here).
Social messages are sent and received by the BDI agents residing in \texttt{Jill
BDI Model}.

\texttt{Messaging Model}: Injects the proposed evacuation plan into the
simulation, which is effectively a series of messages over time sent by the
emergency services. Each message has a type--one of \Advice,
\WatchAndAct, \EmergencyWarning, or \EvacuateNow, some content which
may be interpreted by the receiving agents in some way, and the SA1 zones (see
Figure~\ref{fig:castlemaine-region}) to send the message to. The value of a
message will differ between persons of different archetypes as per
Table~\ref{tab:archetypes-attitudes}. Within persons of the same archetypes,
response to the same message will also differ due to differences in their risk
averseness determined by \ThresholdInitial (see
Section~\ref{subsec:bdi-behaviours}).

\texttt{Data \& Time Control} is the controller that contains the simulation
loop and is responsible for progressing simulation time, calling the components
once in each time step, and managing the data sharing between components. The
data and control scheme is not dislike that prescribed by the
HLA~\cite{dahmann_standards_1998} standard, although, unlike HLA, multiple
models can represent aspects of the same conceptual agents at the same time as
supported by the BDI-ABM integrated framework~\cite{singh_integrating_2016}.
  \section{Evaluation}
\label{sec:evaluation}

\begin{figure}[!t]
  \centering
  \includegraphics[width=0.7\columnwidth]{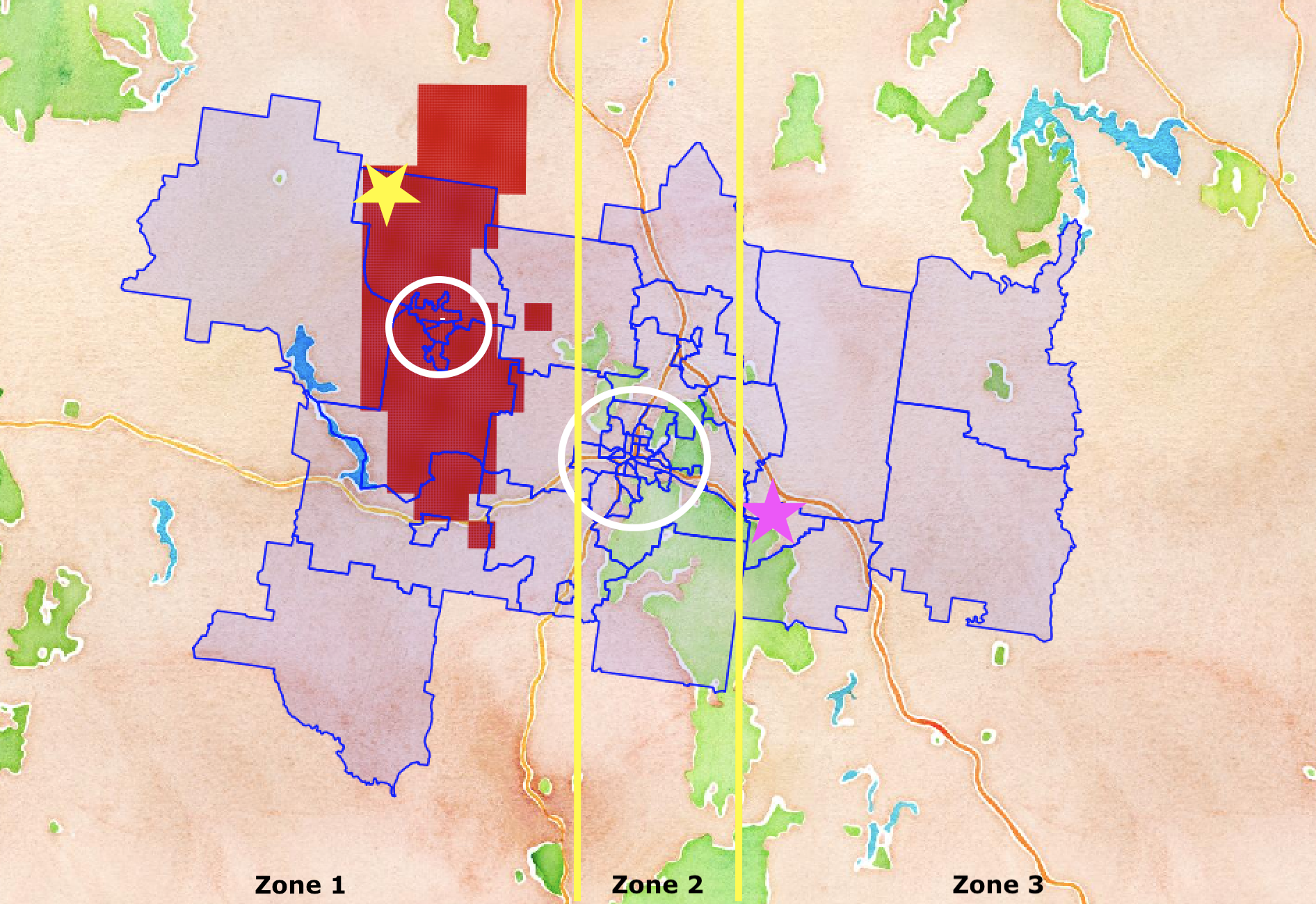}
  \caption{The modelled Castlemaine region (large blue area) lies 120 km
  north-west of Melbourne and constitutes a census (2016) population of 16,623
  persons (total area 1600 km$^2$). Inner blue areas are census level 1
  statistical areas (SA1s) that on average contain 400 people each. The red area
  shows the full extent of the time-varying grid shape of the bushfire generated
  by the Phoenix RapidFire simulator (ignition point shown by yellow star).
  Townships highlighted are Maldon (white circle small) and Castlemaine (white
  circle large). Persons deciding to leave will head in the direction of
  Elphingston (pink star). Zones (separated by yellow lines) are used for
  sending targeted emergency messages to the region.}
  \label{fig:castlemaine-region}
\end{figure}

We applied the method of Section~\ref{sec:method} to build a bushfire evacuation
simulation for the Castlemaine region\footnote{By Castlemaine region, we mean
strictly the combined SA2 areas of Castlemaine and Castlemaine Region in the ABS
2016 census.} shown in Figure~\ref{fig:castlemaine-region}. The region lies 120
km north-west of Melbourne, Australia, and constitutes a census (2016)
population of 16,623 persons.\footnote{As per data downloaded using ABS
TableBuilder.}
The synthetic population created for Castlemaine region (as per
Section~\ref{subsec:population-synthesis}) resulted in 16,489 unique
individuals, to whom we assigned the archetypes (as per
Section~\ref{subsec:archetypes-assignment}).

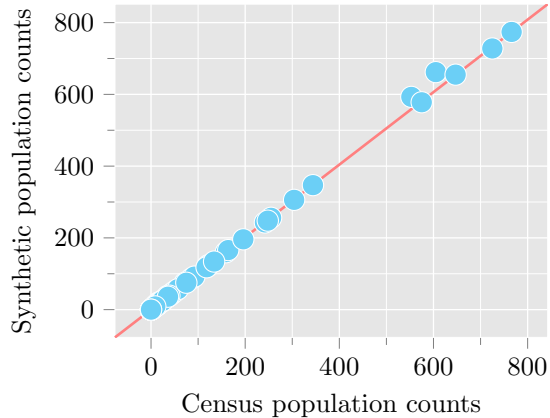
\begin{figure}[!t]
  \centering
  \leavevmode\beginpgfgraphicnamed{archetypes-castlemaine-qqplot-ext}
\pgfplotstableread[col sep =
semicolon]{./data/Castlemaine_Region_persons.csv}\castlereg 

\begin{tikzpicture}

\begin{axis}[axis background/.style={fill=gray!20},
grid=both,
xtick pos=left,
ytick pos=left,
tick style={
	major grid style={style=white,line width=0.5pt},
	minor grid style={style=white,line width=0.25pt},
	tick align=outside,
},
minor tick num=1,
draw=white,
width=0.6\textwidth,
height=6cm,
ylabel={Synthetic population counts},xlabel={Census population counts},]\addplot[
only marks,
fill=cyan!50,
mark size=4pt,
] table [x={Census}, y={Generated}]{\castlereg};
\draw [red!50,line width=1pt] (axis cs:\pgfkeysvalueof{/pgfplots/xmin},\pgfkeysvalueof{/pgfplots/ymin}) --
(axis cs:\pgfkeysvalueof{/pgfplots/xmax},\pgfkeysvalueof{/pgfplots/ymax});

\end{axis}
\end{tikzpicture}
\endpgfgraphicnamed
 
  \caption{Comparison of the Castlemaine region synthetic population (generated
  as per Section~\ref{subsec:population-synthesis}) to the census counts for all
  128 (blue dots) person-level categories.}

	\label{fig:archetypes-castlemaine-qqplot}

\end{figure}

The evaluation showed that the synthetic populations of the two SA2s are very
similar to the corresponding census distributions. At person-level, both SA2s
produced p-values close to 1.0 for the FT test. Thus, none of them can be
concluded dissimilar by rejecting the null hypothesis at the 0.05 significance
level. For the SAE test, Castlemaine and Castlemaine Region SA2s received 0.004
and 0.007, respectively, indicating that the errors are very small.
Figure~\ref{fig:archetypes-castlemaine-qqplot} compares the generated and
census person-level distributions of the two SA2s. Each dot represents a
person-type, and it falls right on the identity ($y = x$) line if the number of
persons in the census (represented by $x$ axis) and generated (represented by
$y$ axis) populations match perfectly. The plots show that all the person-types
are generally very close to the identity line, indicating that generated and
census distributions match reasonably well. Here, we do not discuss
household-level distribution in detail as the Algorithm~\ref{alg:popsynth}
ensures that the generated population is a perfect match to the census
household-level distributions.

\begin{figure*}[!t]
  \centering
  \subfloat[][Castlemaine region assigned archetypes (estimate)]{
    \includegraphics[width=0.49\columnwidth]
    {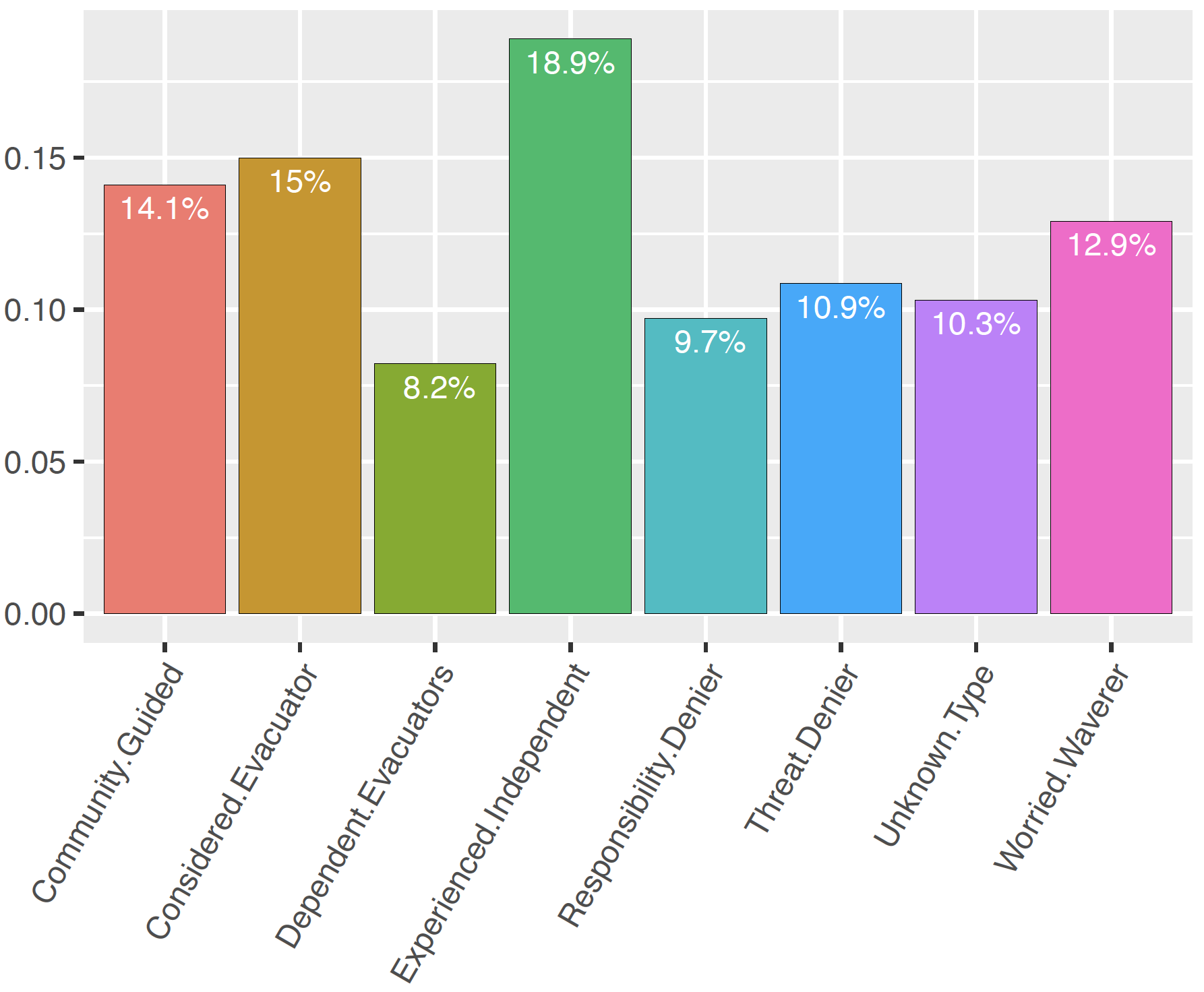}
  }
\subfloat[][Strahan et al. archetypes (actual)]{
    \includegraphics[width=0.49\columnwidth]{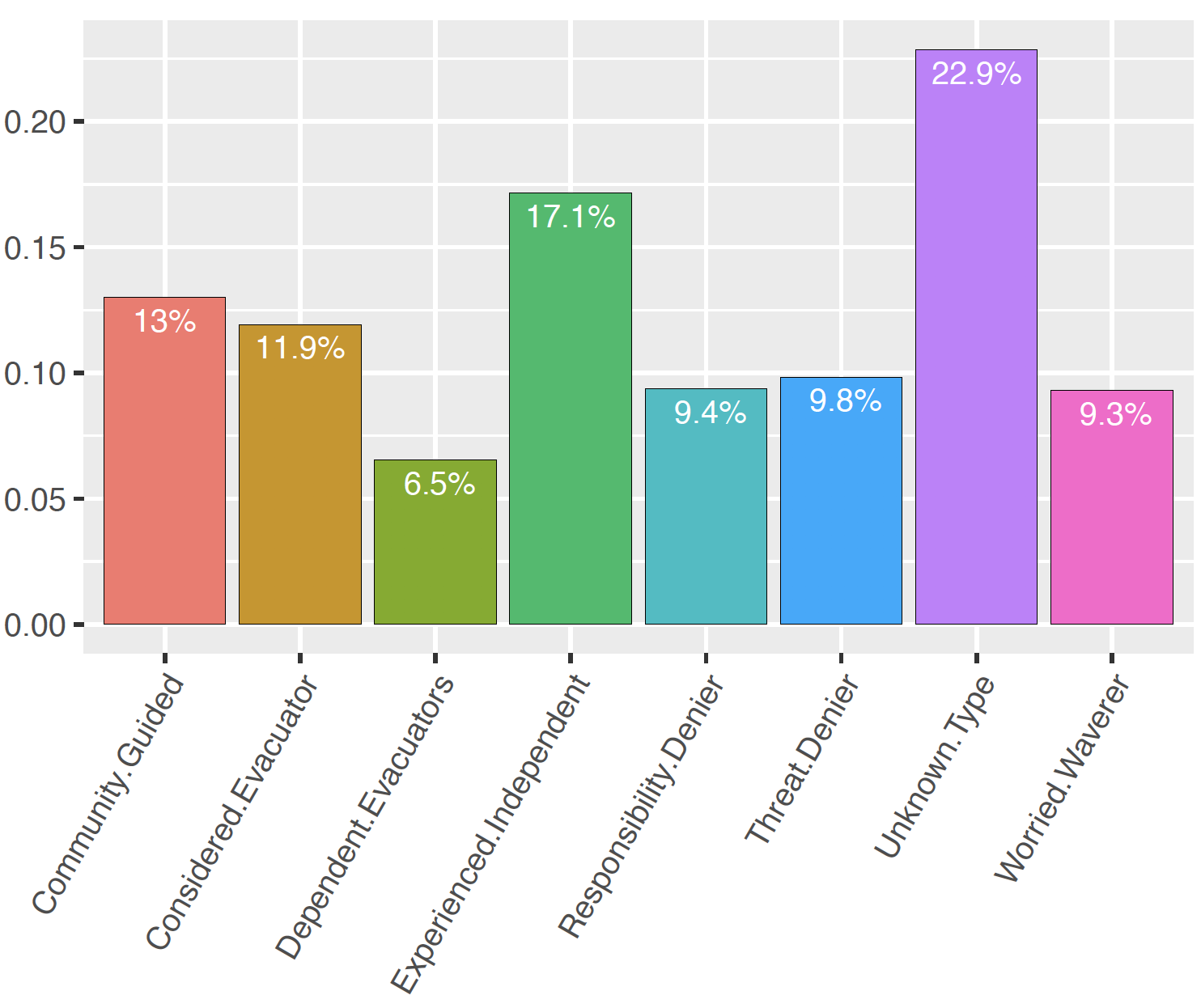}
  }
  \caption{Comparison of the distribution of archetypes assigned to the
  synthetic population of the Castlemaine region (estimate) against Strahan et
  al.'s~\cite{strahan_self-evacuation_2018} data (actual). }
  \label{fig:archetypes-gender-age}
\end{figure*}

Figure~\ref{fig:archetypes-gender-age} shows the makeup of the created
population with respect to the different archetypes, compared against Strahan et
al.'s~\cite{strahan_self-evacuation_2018} original data. The \UT category
includes individuals over 18 years of age for which the archetypes are unknown
(see Section~\ref{subsec:archetypes-probabilities}).
Evidently the archetypal makeup of Castlemaine region is a prediction, purely
based on the demographic profile of the population of the region compared to the
demographic profile of Strahan et al.'s~\cite{strahan_self-evacuation_2018}
original cohort of individuals from the Perth and Adelaide Hills. The natural
question to ask then is, how true is this prediction? To verify this, we are
currently analysing a new data set of post-season telephone surveys conducted by
Strahan for the larger Bendigo SA4 area\footnote{The Castlemaine region
represents 2 of the 16 SA2s that constitute Bendigo SA4.}. Our process is to
construct the full synthetic population of Bendigo SA4, analyse its archetypal
makeup (prediction), and compare it to the archetypal makeup of Bendigo SA4 as
derived from the post-season survey samples (actual). This exercise is work in
progress and is not reported here. The perceived outcome here is a (potentially
revised) robust prediction model of the archetypal makeup of new communities.

For simulation purposes, we removed all the individuals from the Castlemaine
region synthetic population that did not have an archetype assigned,
which included children under the age of 18\footnote{Under 18s
are not part of Strahan et al.'s~\cite{strahan_self-evacuation_2018} original
data and therefore cannot be assigned an archetype.}, any \UT individuals that
did not match the demographic signature of known archetypes (see
Section~\ref{subsec:archetypes-assignment}), and \DE individuals who do not
drive and depend on others to evacuate them to safety.
Dependent children and \DE individuals were instead accounted for by assigning
them as dependants of other individuals in the population (as per
Section~\ref{subsec:archetypes-activities}), so that they get attended to during
the initial response of those individuals to the bushfire (as per
Section~\ref{subsec:bdi-behaviours}).
The final synthetic population used in the simulation
contained 10,868 archetype individuals.

\begin{figure}[!t]
  \centering
  \includegraphics[width=0.7\columnwidth]{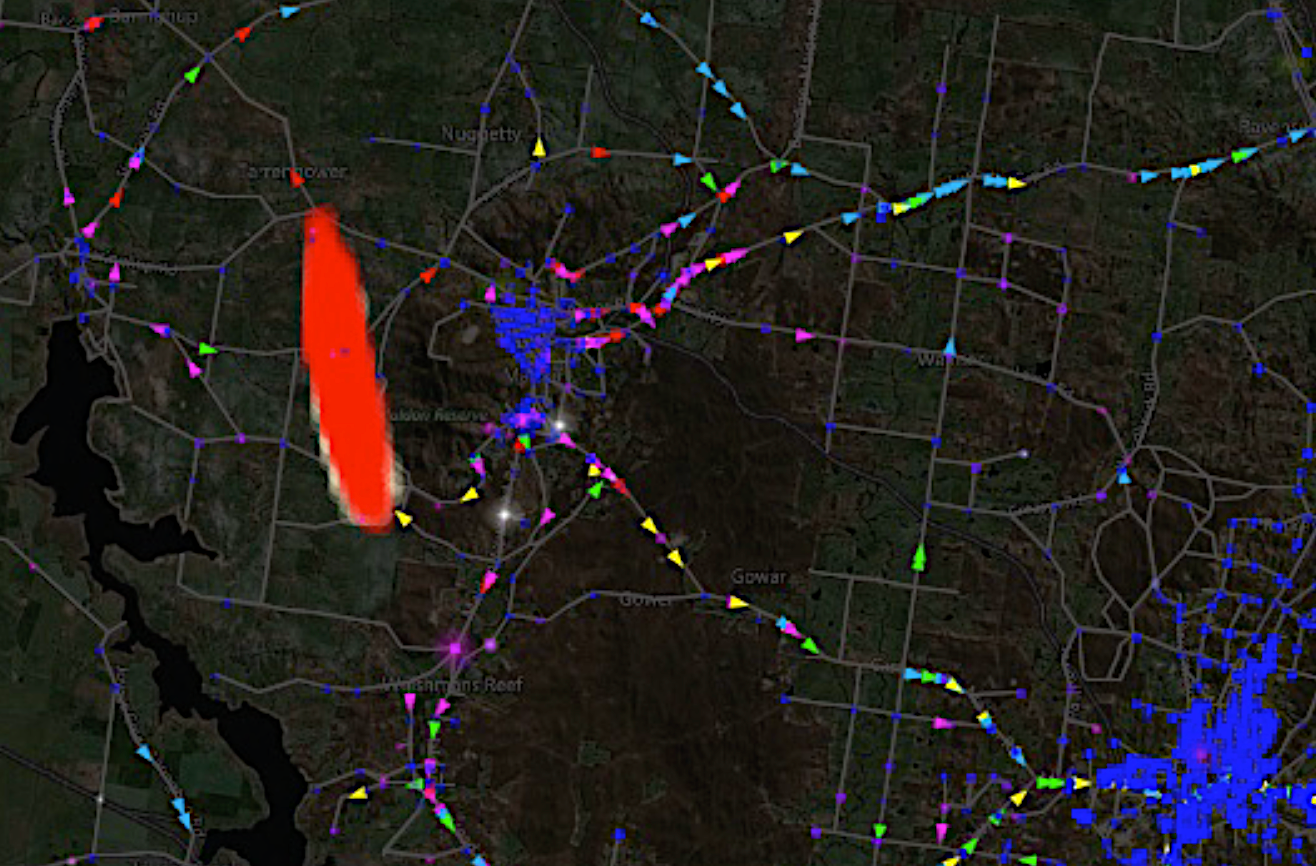}

  \caption{Screenshot of the bushfire evacuation simulation for Castlemaine
  region. The large red patch is the
  progressing fire front, while the light glow around its perimeter is the embers
  front. Blue dots represent people at home, predominantly in the townships of
  Maldon (off centre) and Castlemaine (bottom right). Glowing dots are people
  deliberating over their next action (pink) and people who have just arrived at
  the location of dependants (white). Coloured triangles represent cars on roads
  with different archetype persons, being \WW (blue),  \CG (green), \CE (yellow),
  \RD (pink), \TD (orange), and \EI (red).
  }

  \label{fig:castlemaineRegionArchetypesIT}
\end{figure}

Figure~\ref{fig:castlemaineRegionArchetypesIT} shows a snapshot of the modelled
bushfire evacuation scenario for Castlemaine region.  In this scenario, a
catastrophic fire\footnote{Modelled in Phoenix
RapidFire~\cite{tolhurst_assessing_2011} for weather conditions comparable to a
Code Red day, i.e., a  Forest Fire Danger Index~\cite{dowdy_australian_2009} of
100.} has started to the northwest of Maldon at 11am that eventually engulfs the
town by 5pm (cf. Figure~\ref{fig:castlemaine-region}). Based on discussions
with emergency services, the emergency messages regime chosen for this situation
is as shown in Table~\ref{tab:messages-regime}.

\begin{table}[!t]
\centering
\caption{Emergency messages sent in the Castlemaine region bushfire scenario.}
\label{tab:messages-regime}
\begin{tabular}{|l|p{0.15\textwidth}|p{0.45\textwidth}|}
\hline
{\bf Message} & {\bf Time sent} &
{\bf Region sent to (Figure~\ref{fig:castlemaine-region})}\\
\hline
\hline
\Advice           & 1100 hrs & Zone1 (western region)\\
\WatchAndAct      & 1200 hrs & Zone1 (western region)\\
\EmergencyWarning & 1230 hrs & Zone1 (western region)\\
\EvacuateNow      & 1300 hrs & Zone1 (western region)\\
\Advice           & 1330 hrs & Zone2 (inner region)\\
\EmergencyWarning & 1500 hrs & Zone2a (SW part of Castlemaine circle) \\
\WatchAndAct      & 1600 hrs & Zone2 (inner region)\\
\hline
\end{tabular}
\end{table}

Since data sources on the actual movements of people in bushfires are not
readily available, it is difficult to validate the trajectories of vehicles
observed in the simulation. However, since the 2009 Black Saturday fires in
Victoria, significant work has been conducted in surveys of affected people, and
it is much easier to find anecdotal accounts of what people did in such
situations when on the road~\cite{matthews_capturing_2015,mclennan_predictors_2014,mclennan_householder_2012,strahan_predicting_2019,mccaffrey_should_2018}.
Our process for calibrating and validating the simulation model has therefore
been one of randomly sampling individuals and visually inspecting their trip
chains and routes in MATSim (Figure~\ref{fig:castlemaineRegionArchetypesIT})
together with the output log of their BDI reasoning that shows when they chose
to do what and for what purpose. This allows us to build \textit{virtual}
anecdotal accounts of these individuals, not too dissimilar from those
constructed from post-event surveys of real individuals.
These virtual accounts were then compared to Strahan et
al.'s~\cite{strahan_self-evacuation_2018} understanding of how those archetypal
individuals are likely to behave. If we found a discrepancy, we revisited the
BDI behaviour model (Section~\ref{subsec:bdi-behaviours}) or attitudes
assignment (Section~\ref{subsec:archetypes-attitudes}) and made adjustments as
necessary. We stopped when our random sampling no longer produced virtual
accounts that were inconsistent with our understanding of the archetypes.

Archetypes behaved in the virtual accounts, at a macro level, as the researchers
and domain experts consulted had expected. For example, \CE agents tended to act
early in the bushfire event, responding to warnings from the emergency services
and their own assessment of the immediacy of the threat based on ember and flame
cues. \CE and \CG represented many individuals who acted early by picking up
dependants (if they had them) and deciding to return home or driving directly to
an evacuation centre. They appeared in the simulation visualisation as streams
of yellow (\CE) and green (\CG) agents moving away from the fire front toward
the evacuation centre in the first few hours of the event. On the other hand,
few \EI or \TD agents appeared on the roads early in the event. They stayed to
defend or refused to see the fire as a threat. They were expected not to respond
to warnings and to leave at the last minute when either their defensive efforts
failed or when the threat became too immediate to continue to deny. \EI agents
did not begin to move until later in the fire and \TD even later when a wind
change moved the fire front and blew embers over them substantially heightening
the level and immediacy of the threat.

The manual inspection of select individual trip chains and associated decision
making supported the observations of archetypal evacuation behaviour at the
micro level. For example, a \CE agent left very early in the fire event, to pick
up a dependent, then returned home (perhaps to pick up other family members, a
pet or personal belongings) before leaving again almost immediately for the
evacuation centre. Other agents representing various archetypes, located above
the fire, with it moving away from them, did not receive warnings and until a
wind change moved embers and flames in their direction, remained at their
residences. Some of these were \CE and \CG types who tend to leave early. Such
delayed evacuation for these types reflects what might be expected in reality
when there is no official warning, as they carefully monitor the fire, assess
its likely impact and respond to the conditions and increasing threat created by
the change in wind direction, blowing the fire toward them.

Some individual trip chains highlighted areas in which further work on the model
is required. For example, a \RD acting consistent with their view that others
should take responsibility for dealing with the fire did not respond until
warnings had escalated to the highest levels. Only at that point did they
decide to act, proceeding to pick up a dependent who was incidentally located
within the fire zone and then returning home. This behaviour was deemed unlikely
since the knowledge of dependants in the danger zone should have triggered a
decision to pick them up early not late. Another unlikely behaviour was when an
agent who had no dependants chose to go home to ‘wait and see’, however their
house was located close to the fire zone. In this case the individual should
either have waited where they originally were or evacuated immediately which
would both have been safer options. These kinds of specific accounts give useful
insight into our own understanding of behaviours that may have been
oversimplified, overlooked, or misunderstood, and can be addressed through
further refinement of the model.

Overall, unlikely behaviours were not sufficiently widespread to warrant further
revision of the model.
On the whole the model exhibited a variety of known evacuation behaviours,
including: agents initially going to pick up dependants, (visiting or not their
homes in the process); waiting to see how the situation unfolded (at locations
where they initially were, or at home, or at the dependants location); and
eventually evacuating (via their homes if deemed necessary) to the recommended
evacuation location outside the region (but falling back to a preferred
within-region location or their home location if the attempt to evacuate the
region failed due to roads being cut off by the fire or heavy congestion on
evacuation routes).
 
Our early findings indicate that Strahan et
al.'s~\cite{strahan_self-evacuation_2018} archetypes are robust and hold well
across communities. Were this the case, predictions such as those in
Figure~\ref{fig:archetypes-gender-age} can already be very valuable for the
community. For instance, in the Castlemaine region, a community education program
could be tailored specifically for the predicted large cohort of \EI individuals
who are likely to stay and defend their property at an increased risk to their
lives. Videos of simulation runs against a range of weather and bushfire
conditions can help to inform communities of potential gaps in their
understanding of bushfire risk to them. For instance, in the simulation run
shown in Figure~\ref{fig:castlemaineRegionArchetypesIT} we saw a late out flux
of people from Maldon around 5pm when, due to a wind change, the bushfire started
to encroach on the town. On inspection we found that many of those who left late
from Maldon were \EI individuals.

In our current work in the Castlemaine region,
we are exploring--together with the communities and emergency services--how the
use of this evacuation modelling technology can help with community planning and
preparedness.
 \section{Discussion and Conclusion}
\label{sec:discussion}

In this paper we showed how Strahan et al.'s~\cite{strahan_self-evacuation_2018}
archetypes have been used to build a behaviour model that captures the kinds of
attitudes seen in communities in the event of major bushfires. The model
combines the fire progression output--a time-varying fire front--generated from
the Phoenix RapidFire model used currently by the emergency services to predict
fire behaviour, with the MATSim traffic simulator and a BDI-based behaviour
model to simulate movements of the population of a region in response to that
fire. The outputs of the simulation including visualisations of traffic are used
by the emergency services to improve understanding of community risks and to
improve planning and preparedness. The system is initially being trialled in
Mount Alexander Shire, Surf Coast Shire, Colac Otway Shire, and Yarra Ranges
Shire that together constitute an area of 9,000 km$^2$.

While there have been several efforts since the 2009 Black Saturday fires to
understand the decision making of residents in terms of whether they will stay
and defend or leave~\cite{mclennan_predictors_2014,mclennan_householder_2012,strahan_predicting_2019,mccaffrey_should_2018},
data on what actually happens
during the process of evacuation is scarce. This is because
data collection--such as of movements--during an unfolding
bushfire event, is made difficult by the nature of the emergency.
As such, understanding of behaviours during evacuations has to date been limited
to post-event surveys and accounts~\cite{whittaker_community_2013,handmer_human_2011,handmer_examining_2016,mclennan_at-risk_2015}.

Our approach to modelling evacuation behaviour has been one of collaboration
with bushfire experts and emergency services personnel eliciting knowledge from
their lived experience as well as from existing reports and research papers. It
is worth mentioning that the notion of ``validating'' such a model is already
elusive. For what does it mean to say that a computer model of human
behaviour--one that if not random is inherently tied in rationality--for
life-threatening emergencies where high levels of anxiety can distort rational
thinking, has been validated to be correct? Even if one could make the claim,
how is one to convince others of it? This has indeed been our own experience of
several years.
What has consistently worked however, is a process of co-creation and
iterative refinement of such a model with domain experts.
Key in this has been to ensure that the model exhibits in sufficient detail the
diversity of behaviour seen in actual bushfires, while still being general
enough to be applied to different regions.
Importantly the process has led to stakeholders understanding and owning the
model's assumptions, limitations, and approximations, which have come about
after robust discussions between experts with potentially
differing view points.
The result is a model that is informally validated in the sense that it is
consistent with the understanding of the various experts and exhibits the kinds
of behaviour traces that match anecdotal accounts from survivors. What adds some
sense of trust in the use of such a model for decision making is the fact that
behaviour parameters can be varied over ranges of values (where calibration is
not possible), such that the results can be intuitively interpreted as likely,
best, or worst case outcomes.

Our current work is focusing on improving the answer to ``what are they doing at
the time of the bushfire?'' (see discussion of issues in
Section~\ref{subsec:archetypes-activities}), given that evacuation outcomes are
sensitive to locations of agents. It seems unlikely that the solution will be an
algorithm that generates these populations on demand even if such an algorithm
is conceivable, since the use of any population requires careful thought and
consideration of the underlying assumptions. What seems more likely is that
libraries of populations--such as for weekdays, weekends, and special
events--will be built, critiqued, and approved for operational use.

A remaining big challenge is to move from individuals to households. Even though
we already have the model of household structures now (see
Section~\ref{subsec:population-synthesis}), the big unknown is our understanding
of the dynamics of interactions between household members. For instance, how are
tasks like picking up dependants and going home to collect items shared between
household members, given their locations and available resources
(cars per household)? What does decision making look like for a household where
one member is a \TD and another a \WW? We intend to begin analysing post-season
telephone surveys in Victoria conducted by Strahan over the past two years
($n\approx2000$ to date) where a subset of interviews are also household
members. This will be a start at gaining understanding of the dynamics of
archetypal households.
 \section{Acknowledgements}We thank (in alphabetical order)
Country Fire Authority (CFA),
Commonwealth Scientific and Industrial Research Organisation (CSIRO) Data 61,
Emergency Management Victoria (EMV),
Department of Environment, Land, Water and Planning (DELWP),
Department of Premier and Cabinet Victoria (DPC),
Mount Alexander Shire Council,
RMIT University.
Safer Together program,
Surf Coast Shire Council,
VicRoads,
Victoria Police, and
Yarra Ranges Shire Council
for ongoing support for the bushfire evacuation modelling project.
We thank (in alphabetical order)
Jason Amos (Mount Alexander Shire Council),
Peter Ashton (Surf Coast Shire Council),
Raphaele Blanchi (CSIRO),
Chaminda Bulumulla (RMIT University),
Steve Cameron (EMV),
Trevor Dess (DELWP),
Callum Fairnie (Colac Otway Shire Council),
Tim Gazzard (DELWP),
Justin Halliday (DPC),
Vincent Lemiale (CSIRO Data61),
Leorey Marquez (CSIRO Data61),
John Gilbert (CFA),
Kai Nagel (TU Berlin),
Lin Padgham (RMIT University),
Mahesh Prakash (CSIRO Data61),
Luke Ryan (Mount Alexander Shire Council),
Rajesh Subramanian (CSIRO Data61), and
for their immensely valuable contributions to the work presented in this paper.
 \newpage
\appendix

\section{Castlemaine region simulation video}

A video of the Castlemaine region simulation of Section~\ref{sec:evaluation} is available at \url{https://cloudstor.aarnet.edu.au/plus/s/sG6VjNSmHGjHB3c}.

\section{Summary of data sources and tools used}

\begin{longtable}{|p{0.35\textwidth}|p{0.65\textwidth}|}
\hline
{\bf Task} & {\bf Data Sources and Tools}\\
\hline
\hline
~&~\\
\ref{subsec:archetypes-probabilities}~Estimating the likelihood of archetypes &
Strahan et al's.~\cite{strahan_self-evacuation_2018} raw data was used to
estimate the likelihood of archetypes; this data however has not been made
publicly available at this stage due to privacy contraints.
\\~&~\\
\ref{subsec:population-synthesis}~Synthesising the population for a region
(Algorithm~\ref{alg:popsynth}) & The synthetic population was produced using the
algorithm available from \url{https://github.com/agentsoz/synthetic-population};
the input 2016 census data was sourced from Australian Bureau of Statistics and
available from \url{https://abs.gov.au/census}.
\\~&~\\
\ref{subsec:archetypes-assignment}~Assigning archetypes to the population
(Algorithm~\ref{alg:archetypes-assignment}),
\ref{subsec:archetypes-attitudes}~Assigning attitudes to archetypes,
\ref{subsec:attitudes-calibration}~Calibrating attitudes of archetypes
(Algorithm~\ref{alg:attitudes-calibration}) & All R code and related scripts
used for these tasks are available from
\url{https://bitbucket.org/dhixsingh/archetypes-modelling}, however the
repository has currently not been made public to preserve privacy of
Strahan et al's.~\cite{strahan_self-evacuation_2018} raw data.
\\~&~\\
\ref{subsec:archetypes-activities}~Assigning activities to the population &
Expert opinion based activities assignment algorithm used for Surf Coast Shire
is available from \url{https://github.com/agentsoz/ees-synthetic-population}
(though not used here); Address Mapper algorithm for home locations assignment
for the Castlemaine region simulation is available from
\url{https://github.com/agentsoz/synthetic-population}, where the legal street
addresses for the region were sourced from LandVic and available from
\url{http://services.land.vic.gov.au/landchannel/content/productCaalogue}.

\\~&~\\
\ref{subsec:bdi-behaviours}~Assigning behaviours to the population & The
evacuation behaviour model that determines what agents do once they decide to
evacuate is written in the Jill BDI language available at
\url{https://github.com/agentsoz/jill}.
\\~&~\\
\ref{subsec:ees}~Putting it all together & The road network was extracted from
OpenStreetMap at \url{https://www.openstreetmap.org}, bushfire progression model
outputs from Phoenix RapidFire were provided by Department of Environment, Land,
Water and Planning (DELWP), the full Emergency Evacuation Simulator (EES) model
and Castlemaine region scenario are available from
\url{https://github.com/agentsoz/ees}, with the various components being the
BDI-ABM integration framework from
\url{https://github.com/agentsoz/bdi-abm-integration}, the MATSim traffic
simulator from \url{https://github.com/matsim-org/matsim}, and the social
information diffusion model from
\url{https://github.com/agentsoz/diffusion-model}.
\\~&~\\
\hline
\end{longtable}
 \bibliographystyle{unsrt}
\bibliography{main,popsynth}

\end{document}